%% file: main.tex
\documentclass{article}

\PassOptionsToPackage{textsize=tiny}{todonotes}

\usepackage{microtype,subcaption, bm}
\usepackage{graphicx, multirow,colortbl,hhline,cite}
\usepackage{amsmath}

\usepackage{booktabs} 
\usepackage{xcolor}
\definecolor{tbl1}{rgb}{1,1,1} 
\definecolor{tbl2}{rgb}{1,1,1} 
\usepackage{changes}
\usepackage{longtable}

\usepackage{dsfont, placeins}
\newcommand{\Xset}{\mathcal{X}}
\newcommand{\Yset}{\mathcal{Y}}
\newcommand{\Prob}{\mathbb{P}}
\newcommand{\Real}{\mathbb{R}}
\newcommand{\Dset}{\mathcal{D}}
\newcommand{\Ex}{\mathbb{E}}

\newcommand{\eps}{\epsilon}
\newcommand{\xvec}{\mathbf{x}}
\newcommand{\Xvec}{\mathbf{X}}
\newcommand{\yvec}{\mathbf{y}}
\newcommand{\Yvec}{\mathbf{Y}}

\newcommand{\bigone}{\mathds{1}}
\definecolor{customgreen}{rgb}{0.33, 0.65, 0.18}

\definechangesauthor[name=Marco, color=customgreen]{MR}

\definechangesauthor[name=Georg, color=purple]{GP}

\DeclareMathOperator*{\argmin}{arg\,min}

\usepackage{hyperref}

 \usepackage[accepted]{style2024}

\usepackage{amsmath}
\usepackage{amssymb}
\usepackage{mathtools}
\usepackage{amsthm}

\usepackage[capitalize,noabbrev]{cleveref}

\theoremstyle{plain}
\newtheorem{theorem}{Theorem}[section]
\newtheorem{proposition}[theorem]{Proposition}

\theoremstyle{definition}
\newtheorem{definition}[theorem]{Definition}

\theoremstyle{remark}

\crefformat{equation}{#2(#1)#3}
\crefrangeformat{equation}{#3(#1)#4--#5(#2)#6}
\crefmultiformat{equation}{#2(#1)#3}{ and~#2(#1)#3}{, #2(#1)#3}{ and~#2(#1)#3}
\Crefformat{equation}{#2(#1)#3}
\Crefrangeformat{equation}{#3(#1)#4--#5(#2)#6}
\Crefmultiformat{equation}{#2(#1)#3}{ and~#2(#1)#3}{, #2(#1)#3}{ and~#2(#1)#3}

\titlerunning{Beyond the Norms: Detecting Prediction Errors in Regression Models}

\begin{document}

\twocolumn[
\title{Beyond the Norms: Detecting Prediction Errors in Regression Models}

\confsetsymbol{equal}{*}

\begin{authorlist}
\confauthor{Andres Altieri}{L2S}
\confauthor{Marco Romanelli}{NYU}
\confauthor{Georg Pichler}{TUWien}
\confauthor{Florence Alberge}{SATIE}
\confauthor{Pablo Piantanida}{ILLS}
\end{authorlist}

\confaffiliation{L2S}{Laboratoire des signaux et systèmes (L2S), Université Paris-Saclay CNRS CentraleSupélec, Gif-sur-Yvette, France}
\confaffiliation{NYU}{New York University, New York, NY, USA}
\confaffiliation{TUWien}{Institute of Telecommunications, TU Wien, Vienna, Austria}
\confaffiliation{ILLS}{International Laboratory on Learning Systems (ILLS) and Quebec AI Institute (Mila), McGill ETS CNRS  Université Paris-Saclay CentraleSupélec, Montreal (QC), Canada}
\confaffiliation{SATIE}{Systèmes et applications des technologies de l'information et de l'énergie (SATIE), CNRS  Université Paris-Saclay, Gif-sur-Yvette, France}

\confcorrespondingauthor{Andres Altieri}{andres.altieri@centralesupelec.fr}
\confcorrespondingauthor{Marco Romanelli}{mr6852@nyu.edu}

\vskip 0.3in
]

\printAffiliationsAndNotice{} 

\begin{abstract}
This paper tackles the challenge of detecting unreliable behavior in regression algorithms, which may arise from intrinsic variability (e.g., aleatoric uncertainty) or modeling errors (e.g., model uncertainty). First, we formally introduce the notion of \emph{unreliability in regression}, i.e., when the output of the regressor exceeds a specified discrepancy (or error).
Then, using powerful tools for probabilistic modeling, we estimate the discrepancy density, and we measure its statistical diversity using our proposed metric for \emph{statistical dissimilarity}. In turn, this allows us to derive a data-driven score that expresses the uncertainty of the regression outcome.
We show empirical improvements in error detection for multiple regression tasks, consistently outperforming popular baseline approaches, and contributing to the broader field of uncertainty quantification and safe machine learning systems. 
Our code is available at \url{ https://zenodo.org/records/11281964}.
\end{abstract}

\section{Introduction}
In recent years, machine learning has gained ground in fields such as automatic processing, and autonomous decision-making, generating a crucial need for operational safety and reliability, aiming to prevent catastrophic errors~\cite{amodei_concrete_2016,10.1145/3555803,pnas.1907377117}. The need for trustworthy models has generated a lot of research in areas concerned with the identification of anomalous patterns that may trigger undesired, and potentially dangerous, predictions such as out-of-distribution detection \cite{gomes2022igeood,ming2023how}, selective classification \cite{CorbiereTBCP2019NeurIPS,HuangZZ2020NeurIPS}, misclassification detection \cite{GraneseRGPP2021NeurIPS}, and adversarial attacks  \cite{goodfellow2015explaining, 9773978}, among others.

Anomalies may stem from intrinsic variability in the data, i.e., \emph{aleatoric uncertainty} \cite{KIUREGHIAN2009105}. For instance, certain inputs may exhibit a larger dispersion of the dependent variable than expected, i.e., \emph{heteroscedastic uncertainty} \cite{Kendall2017}. Additional sources of unreliable behavior may be associated with approximation errors or model uncertainty~\cite{KIUREGHIAN2009105}, occurring when the regressor is not statistically close to potential observations. This discrepancy may arise from various factors such as sub-optimal model selection, inadequate model fitting, and insufficient training, among others.

The challenge of identifying anomalous predictions in machine learning models is closely connected to the task of uncertainty quantification, which aims to assess the uncertainty inherent in model predictions \cite{DBLP:conf/nips/KotelevskiiAFNF22,DBLP:conf/nips/LiuLPTBL20}. 
In classification tasks, the process typically involves estimating a probability distribution across labels, introducing a clear distinction between correct and incorrect decisions and a straightforward understanding of uncertainty. In stark contrast, regression problems often present a more convoluted landscape: 
a single regressor output is not sufficient to evaluate the model uncertainty. Thus, an additional uncertainty estimation in the form of a conditional distribution or prediction interval is usually incorporated. 


This work aims to develop a simple yet effective framework to evaluate the reliability of the predictions of a given regressor and  detect potentially anomalous situations that may arise. Our approach involves data-driven techniques to address the inherent challenges in the detection problem, particularly in compensating for estimation inaccuracies associated with baseline methods relying on the estimation of the conditional distribution of the dependent target variable.

\subsection{Contributions} 
Our main contributions can be summarized  as follows:
\begin{enumerate}
\item We study a novel detection challenge focused on identifying inaccuracies in the predictions made by a regressor. Our definition of ``incorrect'' predictions 
references instances in which the regressor output may deviate beyond a predefined threshold from typical observations. 
\item We formalize two fundamental base predictors that serve as benchmark models for the broader detection problem. These are constructed through approximation of the optimal detector and rely on estimates of the conditional distributions of the target and error variables.
\item We utilize an empirical measure of diversity inspired by~\cite{RAO1982} to develop two novel data-driven algorithms. Crucially, the advantage rooted in their inherent capability to compensate for inaccuracies in the underlying probability estimates, allows for them to achieve superior performance in detecting incorrect predictions of regressor models across various datasets and different error metrics.

\end{enumerate}

\subsection{Related Work}

Over recent years, a multitude of methods have been introduced aiming to quantify the uncertainty in model predictions, each grounded in distinct underlying hypotheses, and generally producing an output conditional distribution or prediction interval. Noteworthy examples encompass techniques based on the assumptions of both parametric  models, where pertinent parameters are learned from the data~\cite{Lakshminarayanan17,Rio,KNIFE_2022}, Bayesian approaches~\cite{hernandez-lobatoc15, SharmaAzizanEtAl2021,ritter2018a}, non-parametric quantile function estimates~\cite{Fasiolo2021, Chung2021_Cali, Tagasovska_2019}, regression with reject option~\cite{10.5555/3495724.3497409}, estimation of conditional variance~\cite{6beb6e26-9d87-3973-947c-6322c6dba60d}, confidence intervals~\cite{pmlr-v80-pearce18a}, and conformal intervals~\cite{Papadopoulos2002, Romano2019}. Additionally, a related challenge lies in gauging the accuracy of these predictions, involving concepts such as calibration and sharpness~\cite{Zhao2020,Kuleshov2018AccurateUF}, among others.

The challenges of detecting anomalous behavior in regression algorithms and computing a conditional estimate for dependent variables are related, yet not equivalent. Constructing a detector does not necessarily demand a comprehensive and accurate statistical characterization of dependent variables. Instead, it requires an approach that excels in the specified detection task, capable of extracting relevant information even from suboptimal estimates. To draw an analogy, in classification problems, obtaining well-calibrated estimates of class probabilities~\cite{guo_calibration_2017} aids in detecting classification errors or out-of-distribution samples. However, the performance of such detectors can be enhanced through pre-processing techniques and constructing statistics that capture the pertinent information needed to detect specific conditions~\cite{ODIN_2018, GraneseRGPP2021NeurIPS, gomes2023a}.

\begin{figure}[t!]
\begin{subfigure}[t!]{0.49\linewidth}
\includegraphics[width=1\columnwidth,trim=.5cm 0 0.45cm 0 ,clip ]{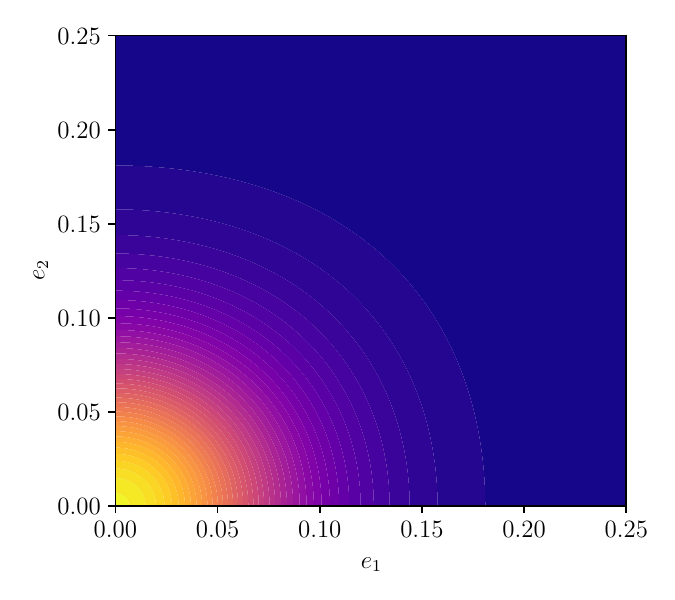}
\caption{Theoretical average joint error distribution for reliable regressor inputs.}
\end{subfigure}\hfill
\begin{subfigure}[t!]{0.49\linewidth}
\includegraphics[width=1\columnwidth,trim=.5cm 0 0.45cm 0,clip ]{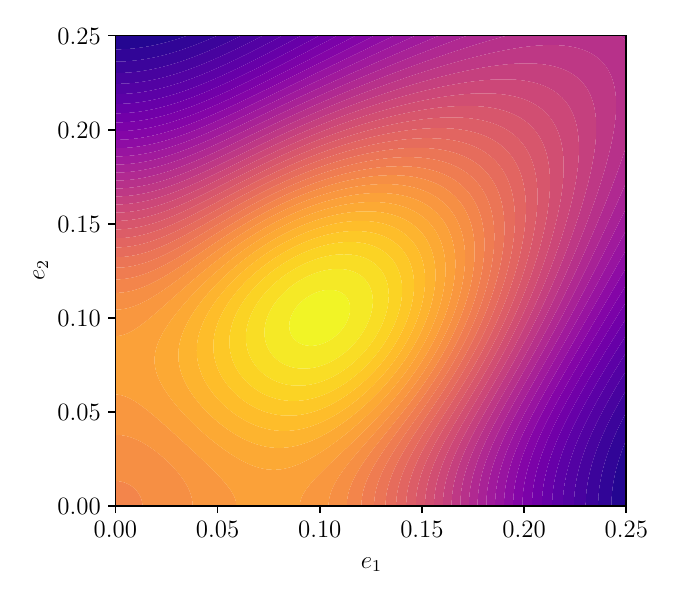}
\caption{
Theoretical average joint error distribution for unreliable regressor inputs.
}
\end{subfigure}
\vfill
\begin{subfigure}[t!]{0.49\linewidth}
\includegraphics[width=1\columnwidth,trim=.3cm 0 0cm 0 ,clip ]{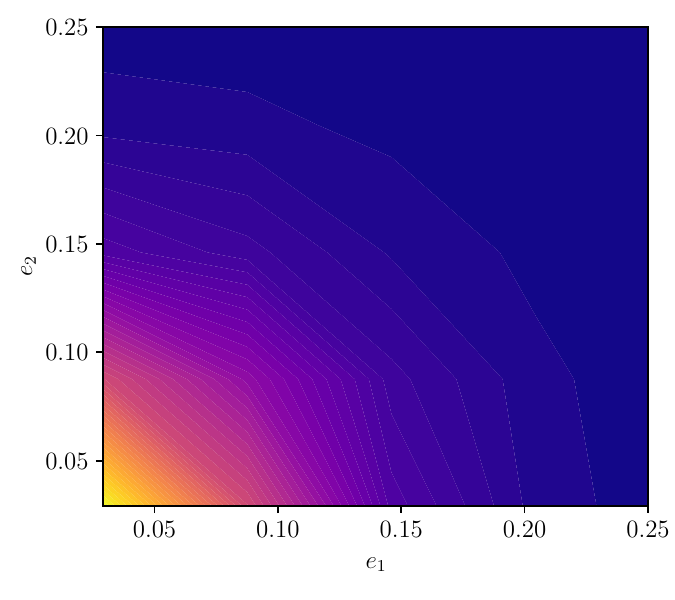}
\caption{Regions from (a) but estimated from data using  SQR.}
\end{subfigure}\hfill
\begin{subfigure}[t!]{0.49\linewidth}
\includegraphics[width=1\columnwidth,trim=.4cm 0 0.3cm 0,clip ]{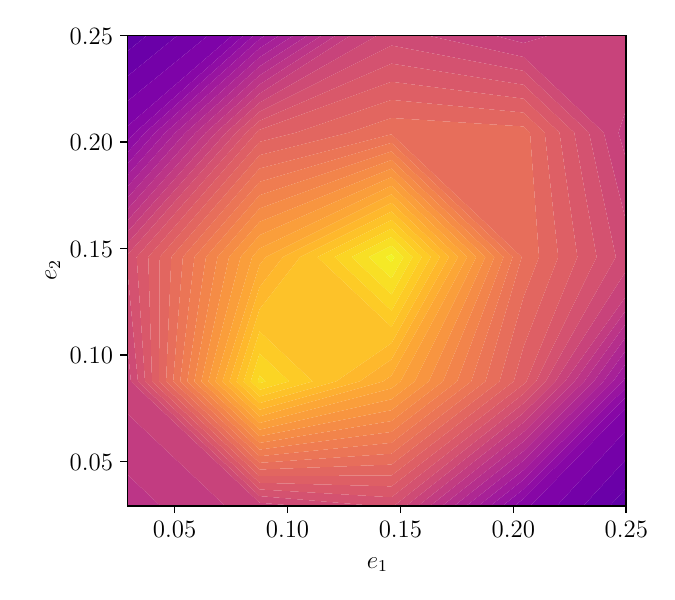}
\caption{Regions from (b) but estimated from data using  SQR.}
\end{subfigure}
\caption{An optimal detector for regression errors effectively partitions the regressor input space into two sets: one where the regressor is considered reliable, and another where it is considered unreliable. Surprisingly, the joint error distributions in each set exhibit distinct (diversity) behaviors, enabling the identification of points that lead to effective detectors in real applications. The visual representation above depicts the theoretical distributions in both sets, illustrating clear differences between the two classes. The corresponding approximate distributions learned from the data, shown below, exhibit the same distinctive behavior. These plots correspond to the example in \cref{sec:example}.
}\label{fig:patt_toy}
\end{figure}

\section{Main Definitions and Example} \label{sec:example}
Let $\Xvec$ be a random vector on $\mathcal{X} \subseteq \mathbb{R}^K$ with distribution $p_{\mathbf{X}}$, and let $\Yvec$ be a dependent random vector on $\mathcal{Y} \subseteq \mathbb{R}^M$ with conditional distribution $p_{\Yvec|\Xvec}$. Consider a regressor $f_{\Dset_n}(\xvec)$ that has been trained on a dataset  $\Dset_n = \{(\xvec_i, \yvec_i)\}_{i=1}^n$  drawn independently from the joint distribution $p_{\Xvec \Yvec}$.
Let us consider a scenario where an alert needs to be triggered whenever, for a given input $\mathbf{x}$, there is a significant risk that the output of the regressor exceeds an absolute error threshold $\eps$.  We may define a prediction error random variable $E \triangleq \|\Yvec - f_{\Dset_n}(\Xvec)\|$ and state the condition to be detected as  $E > \eps$. The most powerful detector 
(cf.  \cref{prop:most_powerful_discriminator}) 
would report the regressor as unreliable at $\xvec$, when
\begin{equation}
\Prob\left(E > \eps | \Xvec=\xvec\right) > \gamma, \label{eq:cond}
\end{equation} where $\gamma \in [0, \ 1]$ is a constant design parameter which controls the sensitivity of the detector. The optimal detector splits the input space $\Xset$ into two sets: a set $\Xset_B$ that contains \emph{bad} points, i.e.
\begin{equation}
    \Xset_B \triangleq \{ \xvec : \Prob\left(E > \eps |\Xvec=\xvec\right) > \gamma\},
\end{equation}
    where the decision is that the regressor is unreliable, and a complementary set of good points $\Xset_G = \Xset_B^{\mathsf{c}}$ 
    where the decision is that the regressor is reliable.

\textbf{Remark.} Notice that this problem is not always well-posed: if for a given regressor the probability on the l.h.s.\ of \cref{eq:cond} is independent of $\xvec$, then the optimal criterion becomes trivial. The performance that we can achieve will thus be defined by the dependence of this probability on $\xvec$.

In practice, the optimal detector cannot be implemented since the underlying conditional distribution is not known, so we propose a different approach. For a given input $\xvec$, we consider two independent realizations of the error variables $E_1, E_2$ drawn independently according to the conditional distribution $p_{E|\Xvec = \xvec}$.  The joint distribution of these two variables $p_{E_1, E_2|\Xvec=\xvec}$ is the product distribution:
\begin{equation}
    p_{E_1 E_2|\Xvec=\xvec} (e_1, e_2) = p_{E|\Xvec=\xvec}(e_1) p_{E|\Xvec=\xvec}(e_2). \label{eq:pE1E2}
\end{equation}
The distribution in \cref{eq:pE1E2} is a measure of the relation of the error event over two independent observations.
Our detection approach is based on the hypothesis that the joint error distributions in \cref{eq:pE1E2} are different enough for the values of $\xvec$ in the set of bad points $\Xset_B$ and those in the set of good points. Therefore, by considering these distributions, we can effectively construct a reliable detector.

\paragraph{Example.}
  To showcase the validity of this assumption, consider ${X \sim \mathcal{N}(0,1)}$ and 
  the simple additive noise model 
  \begin{equation}Y = \phi(x) + W,\end{equation} where $\phi(x)$ is a deterministic function and 
  $p_{W|X=x}(w|x) = \mathcal{N}(w ; 0, \sigma^2(x))$.
  The underlying distribution of the difference $Y - f_{\Dset_n}(x)$ given $X=x$ in this  example is: $p_{Y - f_{\Dset_n}(x)|X=x} (e|x) = \mathcal{N} (e ; b(x), \sigma^2(x)), $     
  where $b(x) \triangleq \phi(x)-f_{\Dset_n}(x)$ is the bias error of the regressor. From these  expressions, we can calculate the probability in \cref{eq:cond} and obtain the optimal detector. Utilizing this information, we can determine the set of good and bad points, enabling us to construct the product distribution \cref{eq:pE1E2} for both classes.
  
  In a numerical experiment, we draw 50 realizations of $X$, categorize them into good and bad points, and compute the respective empirical error distributions \cref{eq:pE1E2}. We then average these distributions to represent the average behavior of each class. The results are depicted in Figs.~\ref{fig:patt_toy}(a) and (b) for the good and bad points, respectively.
  These plots illustrate that the product distributions exhibit distinct behaviors for values of $x$ where the regressor would be deemed reliable or unreliable. Thus, these product functions can be employed to construct a discriminator.
  In practice, however, these distributions are unknown and should be estimated from data. In Figs. \ref{fig:patt_toy} (c) and (d), we repeat the experiment for the same samples of $X$, but we estimate the distributions using \emph{Simultaneous Quantile Regression}~\cite{Tagasovska_2019}. As observed, the estimated distributions align well with the theoretical ones, indicating that a discriminator can be effectively constructed from real data\footnote{Additional details on the experiment and results based on  real datasets can be found in Appendix~\ref{app:example}.}. 

\section{Formal Statement of the Problem }
We now establish the formal framework, slightly generalizing the problem introduced in Section \ref{sec:example}. We use the concept of a discrepancy or error function, which quantifies the ``deviation'' of a regressor output $f_{\mathcal{D}_n}(\xvec)$ from the target $\Yvec$. Then, we shall formulate our detection problem.

\subsection{Regression Error Framework}

\begin{definition}[Discrepancy function]
  Let us consider a function $d: \Yset \times \Yset \to \Real_{\geq 0}$  such that, for a pair $(\xvec, \yvec)$, the value $d(\yvec,f_{\Dset_n}(\xvec))$ serves as an error measure of the difference between the  target variable and the prediction.  We call any such function a \emph{discrepancy function}.
\end{definition}
The choice of the discrepancy function depends on the regression task at hand, and could be different according to the application and its requirements. Given a suitable discrepancy function we may now introduce a notion of good or normal behavior for a regressor with respect to this metric.
\begin{definition}[$\eps$-goodness] Given a discrepancy function $d$, an error threshold $\eps > 0$ and a data pair $(\xvec,\yvec)$ we say that the trained regressor $f_{\Dset_n}$ is $\eps$-good with respect to $d$ for the data pair when $d(\yvec, f_{\Dset_n}(\xvec)) \leq \eps$.
\end{definition}

Based on this notion we define a binary random variable which indicates that the regressor is operating beyond the expected limits for a certain data pair:
\begin{equation}
  \label{eq:Aeps}
A_{\eps} \triangleq  \bigone \{ d( \Yvec,f_{\Dset_n}(\Xvec)) \ge  \epsilon\}.
\end{equation}
This variable takes the value $0$ for those data pairs for which the regressor is $\eps$-good and $1$ for those pairs that are $\eps$-bad.
We denote by $P_B(\xvec ; \eps)$ the probability that the regressor is $\eps$-bad given that $\Xvec = \xvec$. This probability can be written as:
\begin{equation}
  P_B(\xvec ; \eps)=  \mathbb{E}\left[A_{\eps}|\Xvec=\xvec\right]. \label{eq:Pbad}
\end{equation}
In a similar fashion we can compute the probability that the regressor is $\eps$-good at $\xvec$, which we denote as $P_G(\xvec;\eps)$ and can be computed as $P_G(\xvec;\eps) = 1 - P_B(\xvec ; \eps)$.

Our goal is to detect inputs for which the regressor is operating beyond specification, that is, inputs for which there is a high risk that the regressor output be $\eps$-bad. The goal is then to design a discriminator to detect this scenario. 

\section{Vanilla Predictor: Baseline}
From a theoretical standpoint, if we are able to compute the exact probability of being $\eps$-bad given by \cref{eq:Pbad}, we could construct the most powerful discriminator as follows.
\subsection{Optimal (oracle) Discriminator} \label{sec:optdect}

\begin{definition}[Most powerful discriminator] \label{def:most_pow} Consider a regressor $f_{\Dset_n}$, a discrepancy function $d$ and an error threshold $\eps$ as in the previous section. Then, for a regressor input $\xvec$  and for any threshold $\gamma \in [0,\ 1]$  define the decision region: 
\begin{equation}
    \mathcal{R}_{\eps}(\gamma) \triangleq \left\{\xvec \in \Xset : P_B(\xvec ; \eps) > \gamma \right\}.\label{eq:opt_reg}
  \end{equation}
\end{definition}

This definition allows us to obtain the Neyman-Pearson most powerful test for $\eps$-goodness. The proof of the following \lcnamecref{prop:most_powerful_discriminator} is presented in \cref{sec:proof-most-powerful}.
\begin{proposition}
  \label{prop:most_powerful_discriminator}
  
  The discriminator $\delta_p(\xvec, \gamma,\eps)$, defined as
    $\delta_p(\xvec, \gamma,\eps)=\bigone\{\xvec \in \mathcal{R}_{\eps}(\gamma)\}$
  is the most powerful statistical test, testing $\eps$-goodness against the alternative that $\xvec$ is $\eps$-bad, at a certain false positive probability, which is parameterized by $\gamma \in [0,1]$.
\end{proposition}

If the distribution of $\Yvec|\Xvec=\xvec$ is known, the probability of being $\eps$-bad can be computed as:
\begin{align}
    P_B(\xvec ; \eps) &= \Ex \left[ \bigone \{ d(\yvec,f_{\Dset_n}(\xvec)) \ge  \eps\}|\Xvec=\xvec \right]\nonumber \\
    & = \int_\Yset \bigone \{d(\yvec,f_{\Dset_n}(\xvec)) \ge \eps\} \ dp_{\Yvec|\Xvec=\xvec}(\yvec). \label{eq:Pg}
\end{align}


Alternatively, we could define a scalar discrepancy random variable $D(\Yvec, \Xvec) \triangleq d(\Yvec,f_{\Dset_n} (\Xvec))$, and compute the probability as follows:
\begin{align}
    P_B(\xvec ; \eps) &= \Ex \left[ \bigone \{ d(\Yvec,f_{\Dset_n}(\xvec)) \ge  \eps\}|\Xvec=\xvec \right]\nonumber \\
    &= 1 - F_{D(\Yvec, \Xvec)|\Xvec=\xvec} (\eps),\label{eq:Pg2}
\end{align}
where $F$ denotes the cumulative distribution function. 


\subsection{Vanilla Baseline Algorithms}\label{sec:baselines}
The most powerful detector offers optimal performance, but  cannot be implemented since the required conditional distributions are unknown in general. However, we can define approximate detectors by using the optimal decision rule with an estimate  $\hat{P}_B(\xvec;\eps)$ of $P_B(\xvec ; \eps)$, that is, implementing an approximate detector:
\begin{equation}
    \hat{\delta}(\xvec, \gamma,\eps) = \bigone \{ \hat{P}_B(\xvec;\eps) > \gamma \}.
\end{equation}
Possible estimates of $P_B(\xvec ; \eps)$ can be obtained from either \cref{eq:Pg} or \cref{eq:Pg2}, by estimating the respective conditional distribution. These two approaches yield the baseline Algorithm \ref{alg:baseline1}, based on estimating the distribution of $\Yvec | \Xvec = \xvec$, and Algorithm \ref{alg:baseline2},  based on the estimation of the distribution of $D(\Yvec, \Xvec)|\Xvec=\xvec$.  Due to the similarities between both algorithms the latter is presented in Appendix \ref{sec:algorithms}. 
\begin{algorithm}[tb]
   \caption{Baseline based on the estimation of the conditional distribution of $\Yvec|\Xvec=\xvec$}
   \label{alg:baseline1}
\begin{algorithmic}
   \STATE {\bfseries Input:} a trained regressor $f_{\Dset_n}$, trained on the dataset $\Dset_n = \{(\xvec_i, \yvec_i)\}_{i=1}^n$. A detection threshold $\gamma \in [0, \ 1]$.
   
\vspace{2mm}
   \STATE \textbf{Training:} Reuse  $\Dset_n$ to estimate  the conditional distribution of $\Yvec|\Xvec$.
   
\vspace{3mm}
\STATE\textbf{After training:} Given an input test sample $\xvec$:

\STATE \textbf{\hspace{3mm} Step 1:} Estimate $\hat{P}_B(\xvec;\eps)$ using \cref{eq:Pg} [involves the regressor $f_{\Dset_n}(\xvec)$]
\STATE \textbf{\hspace{3mm} Step 2:}  Decide that the regressor will be $\eps$-bad at $\xvec$ if $\hat{P}_B(\xvec;\eps) > \gamma$.
\end{algorithmic}
\end{algorithm}
 
It is noteworthy that theoretically, both \cref{eq:Pg,eq:Pg2} are equivalent, as the distribution of $D(\Yvec, \xvec)|\Xvec = \xvec$ can be computed from the distribution of $\Yvec |\Xvec=\xvec$. However, in practice, the estimation problems differ. $D(\Yvec, \Xvec)$ is a scalar non-negative random variable, while $\Yvec$ could be multidimensional and assume arbitrary values.
In Algorithm \ref{alg:baseline1}, the regressor is not explicitly utilized in the training phase of the estimation of the conditional distribution. It is only employed after training to compute the probability $P_B(\xvec ; \eps)$ for a given test sample $\xvec$.  Thus the training of the regressor and the estimator can be done in parallel. In Algorithm \ref{alg:baseline2}, the regressor is employed during training of the estimator of the conditional distribution, but is not used in the operation of the detector in test. In this case the training of the regressor and the estimator has to be done sequentially.





\section{Error Diversity-Based Detectors for Unreliable Regression Behavior}

A limitation of the baseline Algorithms \ref{alg:baseline1} and \ref{alg:baseline2} is that in practice the estimation of the conditional distributions may be difficult, leading to inaccuracies in the estimation of $P_B(\xvec ; \eps)$. To mitigate this issue we propose a new approach  based on the notion of diversity coefficients~\cite{RAO1982}.

\begin{definition}[Diversity coefficients] Consider a population of samples $\pi=\{V_1, V_2,\dots \}$ whose elements are drawn independently from a probability distribution $p_V$ and a non-negative symmetric function $h(v_1, v_2)$ which measures the difference between two elements in the population. The diversity coefficient $\mathbb{H}$ of the population is  defined as the expected difference, as measured by $h$, of two elements of  $\pi$ drawn independently, that is~\cite{RAO1982}:
\begin{equation}
   \mathbb{H} \triangleq \Ex\left[h(V_1, V_2) \right],
\end{equation}
where $V_1$ and $V_2$ are independent,  $V_1\sim p_V$ and $V_2 \sim p_V$. 
This coefficient gives a characterization of the 
variability within a population as measured by the 
chosen function $h$, i.e., the expected difference between 
two independently chosen 
random elements of the same population. To see
this, consider, e.g., $h(u,v) = (u-v)^2$, 
which results in $\mathbb{H} = 2\text{var}(V)$. 
\end{definition}
For each input $\xvec$, we will consider the discrepancy random variable $D=d(\Yvec, f_{\Dset_n}(\xvec))$, where $\Yvec$ is distributed according to $p_{\Yvec|\Xvec = \xvec}$, i.e., conditioned on $\Xvec=\xvec$. For each $\xvec$, we define the diversity coefficient.

\begin{definition}[Diversity metric for regressor]\label{def:div_met} Given a symmetric non-negative bounded function $h(y_1, y_2)$ and a trained regressor $f_{\Dset_n}(\cdot)$ we define the diversity of the discrepancy variables associated to  $\Xvec=\xvec$ as:
\begin{equation}
\mathbb{H}(\xvec) \triangleq \Ex \left[h(d(\Yvec_1, f_{\Dset_n}(\xvec)), d(\Yvec_2, f_{\Dset_n}(\xvec)))|\Xvec=\xvec\right], \label{eq:divC}
\end{equation}
where $\Yvec_1$ and $\Yvec_2$ are drawn independently according to $p_{\Yvec|\Xvec = \xvec}$.
The diversity coefficient $   \mathbb{H}(\xvec)$ can be computed for any input value $\xvec \in \Xset$.
\end{definition}


Given a diversity metric  we could construct a discriminator which separates the points in $\Xset$ according to their diversity. That is, we can implement a discriminator as follows.
\begin{definition}[Diversity discriminator]
Given a diversity metric $\mathbb{H}$ for a regressor, as in Definition \ref{def:div_met}, for a threshold $\gamma>0$ we define a diversity discriminator (DV) as:
\begin{equation}
\delta_\mathbb{H} (\xvec, \gamma) =  \bigone \{    \mathbb{H}(\xvec) > \gamma\}.
\end{equation}
\end{definition}
Changing the function $h$ that defines the diversity metric allows for the creation of different discriminators. Our objective is to implement a diversity discriminator that is sensitive to points where the regressor is $\eps$-bad. In particular, we explore the possibility of implementing the optimal detector presented in Section \ref{sec:optdect} as a diversity discriminator. For this purpose, we present the following proposition. For the proof, refer to Appendix \ref{sec:proof_div}.
\begin{proposition}\label{prop:Hopt}
  The most powerful detector from Definition \ref{def:most_pow} can be implemented as a diversity discriminator, using the non-negative symmetric function
  \begin{equation}
    h_p(u, v)  \triangleq \bigone\{u > \eps\} \bigone\{v > \eps\}. \label{eq:hopt0}
  \end{equation}
\end{proposition}
This proposition indicates that if the true data distributions are known, it is possible to construct the optimal detector using a diversity detector. In real-world scenarios, the true distribution is not known, and thus,  discriminators can only operate on estimates of the distributions. It can be verified that constructing a diversity detector based on these estimates, and using the function \cref{eq:hopt0} of the optimal detector, will recover the baseline algorithms outlined in Section \ref{sec:baselines}. However, since practical situations involve estimates of these distributions, it may be possible to find a different $h$ function that improves the performance of the baselines by compensating for the estimation errors in the distributions.
\begin{algorithm}[t!]
       \caption{\textbf{-- DV-Y}: Diversity discriminator based on the estimates of the distribution of $\Yvec|\Xvec=\xvec$.}
   \label{alg:div_discY}
\begin{algorithmic} 
   \STATE {\bfseries Input:} A regressor $f_{\Dset_n}$ trained on the dataset $\Dset_n$, dissimilarity threshold $\gamma$
\STATE \textbf{Training:}
   \STATE  \textbf{\hspace{3mm}Step 1:} Use  $\Dset_n$ to obtain an estimate $\hat{F}_{\mathbf{Y}|\Xvec=\xvec}$ of  the distribution of $\Yvec|\Xvec$.
\STATE \textbf{\hspace{3mm}Step 2:} separate $\eps$-good and bad samples in $\Dset_n$:
\vspace{-2mm}
\begin{align*}
    \mathcal{G} \triangleq \{ (\xvec, \yvec) \in \Dset_n : d(\yvec, f_{\Dset_n}(\xvec)) \leq \eps\}\\
    \mathcal{B} \triangleq \{ (\xvec, \yvec) \in \Dset_n : d(\yvec, f_{\Dset_n}(\xvec)) > \eps\}
\end{align*}

\STATE \textbf{\hspace{3mm}Step 3:} learn the function $h$ that defined the diversity metric $\mathbb{H}$ for the discriminator:
\FOR {$i$ in $1,..,N_\text{epochs}$} 
\FOR {each $(\xvec, y) \in \Dset_n$}
\STATE Generate two extended datasets of $N_u$ samples:
\vspace{-1mm}
\begin{alignat*}{1}
\Dset_1(\xvec) = \{\bar{\yvec}_1,\dots,\bar{\yvec}_{N_u}\} \hspace{2mm} & \hspace{2mm} \Dset_2(\xvec) = \{\tilde{\yvec}_1,\dots ,\tilde{\yvec}_{N_u}\}.
\end{alignat*}
where each $\bar{y}_i, \tilde{y}_i$ is generated i.i.d.\ with  $\hat{F}_{Y|\Xvec=\xvec}$ from Step 1.
\STATE Estimate the diversity metric for $\xvec$:
$$   \mathbb{H}(\xvec) = \frac{1}{N_u} \sum_{i=1}^{N_u} h(d(\bar{\yvec}_i, f_{\Dset_n}(\xvec)),d(\tilde{\yvec}_i, f_{\Dset_n}(\xvec))),$$
\ENDFOR
\STATE Update $h$ to minimize:
$$\mathcal{L} = \frac{1}{2} \frac{1}{|\mathcal{G}|}\sum_{\xvec \in \mathcal{G}}    \mathbb{H}(\xvec) - \frac{1}{2} \frac{1}{|\mathcal{B}|} \sum_{\xvec \in \mathcal{B}}    \mathbb{H}(\xvec)$$
such that $h$ is symmetric and $h:\mathbb{R}^2_{\geq 0}  \rightarrow [0, 1]$.
\ENDFOR 
\vspace{2mm}

\STATE \textbf{After training:} Given an input test sample $\xvec$:
\STATE \textbf{\hspace{3mm} Step 1:} compute $   \mathbb{H}(\xvec)$ via a Monte Carlo experiment like in Step 3 of training. 
\STATE \textbf{\hspace{3mm} Step 2:} Decide that the regressor will be $\eps$-bad at $\xvec$ if   $\mathbb{H}(\xvec) > \gamma$.
\end{algorithmic}
\end{algorithm}

We propose to use the data to obtain an empirical 
discrepancy metric as an approximate solution to  an optimization problem which induces the desired behavior:
\begin{multline}
h^* \triangleq \argmin_{h:\mathbb{R}^2_{\geq 0}  \rightarrow [0, 1]} \Ex_\Xvec\left[   \mathbb{H}(\Xvec) \bigone\{\Xvec \in \mathcal{R}_{\eps}(\gamma) \}\right] \\- \Ex_\Xvec\left[\mathbb{H}(\Xvec) \bigone\{\Xvec \notin \mathcal{R}_{\eps}(\gamma)\}\right], \label{eq:teo_prob}
\end{multline}
where $\mathbb{H}(\cdot)$ is given by \cref{eq:divC} and $\mathcal{R}_{\eps}(\gamma)$ is the optimal decision region for the ideal detector \cref{eq:opt_reg}. This optimization problem searches for a diversity metric $\mathbb{H}$ which is minimal on the good points and maximal on the bad points, as defined by the optimal detector.
We adapt this optimization problem and derive two algorithms, Algorithms \ref{alg:div_discY} (DV-Y) and  \ref{alg:div_discD} (DV-D). In the former, the expectation that defines  $\mathbb{H}$ is approximated by using the estimates of the conditional distribution of $\Yvec|\Xvec = \xvec$, and in the latter, we use estimates of the conditional distribution of the discrepancy variable  $D(\Yvec, \Xvec)$, also conditioned on $\Xvec = \xvec$. Since the algorithms are similar in nature, DV-D is relegated to the Appendix \ref{sec:algorithms}. The key aspects of the algorithms are as follows:
\begin{itemize}
\item The goal is to learn the symmetric bounded non-negative function  $h$ that is the argument of the expectation in $\mathbb{H}$ (see \cref{eq:divC}) and defines the diversity metric. 
\item The expectation in  $\mathbb{H}$ is estimated through a Monte Carlo experiment using the empirical distribution estimates as a generative model to  sample the data. 
\item The two terms in the loss function in \cref{eq:teo_prob} are  approximated by empirically separating the training set into the samples that happen to be good or bad.
\end{itemize}
Thus, the algorithms will utilize the diversity in the distribution estimates and the empirical information of good/bad points in the training dataset to define a diversity detector which is tailored to the underlying data estimates.



\section{Experimental Results} \label{sec:numerical}

We compare the diversity discriminators DV-Y (Algorithm \ref{alg:div_discY}) and DV-D (Algorithm \ref{alg:div_discD}) with the baseline Algorithms \ref{alg:baseline1} and \ref{alg:baseline2}. 
To do this we require the estimation of the conditional distributions $\Yvec |\Xvec =\xvec$ and  $D(\Yvec, \Xvec)|\Xvec=\xvec$.
To obtain these estimates, we consider several   different distribution estimation methods, namely, \emph{Conditional Gaussian} (CG) \cite{Lakshminarayanan17}, \emph{Simultaneous Quantile Regression} (SQR) \cite{Tagasovska_2019}, KNIFE \cite{KNIFE_2022} and RIO~\cite{Rio}.
CG fits a conditional Gaussian distribution whose mean and variance depend on the input $\xvec$. SQR estimates the conditional quantile function of the data using the \emph{pinball loss}~\cite{pinball78,Koenker_2005} and  KNIFE fits a conditional Gaussian mixture model to the data. Finally, RIO models the errors as a Gaussian processes. Each distribution estimation algorithm yields a different version of the baseline algorithms and diversity detectors. We denote as B1-* and B2-*, where * can be SQR, CG, KNIFE the baseline Algorithms 1 and 2 obtained with the corresponding distribution estimators.  Since RIO models the errors directly, it is only suitable for implementing baseline B2. Likewise we denote as DV-Y-* and DV-D-*, where * can be SQR, CG, KNIFE the corresponding diversity discriminators implemented with the aforementioned distribution estimators.
For fairness, the same estimates will be used for B1 and DV-Y, and for B2 and DV-D. Thus, in our extensive numerical evaluation, we consider six variations of the baselines and six variations of the algorithms based on the diversity metrics (code  as available supplementary data).


\input{tables2/all_abs_red.txt}

\input{tables2/all_rel_red.txt}

\subsection{Setup}
We consider 8 well-known UCI~\cite{UCI} regression datasets  that have been extensively used in  uncertainty quantification~\cite{hernandez-lobatoc15,pmlr-v48-gal16,Chung2021_Cali,Tagasovska_2019}. For each dataset, we consider two discrepancies, namely, the absolute error: $d_a(Y, f_{\Dset_n}(\xvec)) = |y-f_{\Dset_n}(\xvec)|$; and the relative error: $  d_r(y, f_{\Dset_n}(\xvec)) = \frac{|y-f_{\Dset_n}(\xvec))|}{|f_{\Dset_n}(\xvec)|}$,  measuring   relative discrepancy to the amplitude of  predictions.

For each dataset and discrepancy function, we explore three values of $\eps$ and conduct our experiments across 10 different seeds. When considering the relative error, the chosen $\eps$ values signify a relative magnitude, independent of the data range. However, for the absolute error, suitable $\eps$ values may depend on the dataset and could have significantly different ranges. To mitigate this, we report the value of $\eps$ for the absolute error normalized by the empirical standard deviation of the training set. This normalization ensures resulting values of normalized $\eps$ are less than 1. It is important to mention that  the representative values of $\eps$, for the relative and absolute metrics, depend on each specific datasets. It is not possible to use the same values of $\eps$ for all algorithms because each regressor achieves different levels of approximation to the data.  It is also important to mention that for each value of $\eps$ we obtain a fundamentally different detection problem because the rejection region of the optimal detector changes completely. The best performance achievable for a value of $\eps$ depends on the statistical dispersion of the data around the regressor, and on the train/test splits that are drawn from each seed. For this reason the performance within the same value of $\eps$ and the same dataset may change from seed to seed, especially for the smaller datasets.

For each dataset and seed, we initially train a regressor using a consistent neural architecture featuring three hidden layers and 64 neurons. This involves employing 5-fold cross-validation on the learning rate and weight decay, with 90\% of the data serving as the training set and the remaining 10\% as a test set for both the regressors and the detectors. Subsequently, the training set of the regressor is reused to train estimators for the conditional distributions of $Y|\Xvec=\xvec$ and $D|\Xvec=\xvec$. For the mean and variance of CG, we utilize a neural network with a single hidden layer comprising 64 neurons and average over 10 models. The learning rate is optimized through 5-fold cross-validation, and the training continues for 150 epochs. Adhering to the setup in \cite{Tagasovska_2019}, SQR is trained using a neural network with 3 hidden layers and 64 neurons. The learning rate and weight decay are optimized through 5-fold cross-validation, and the training proceeds for 1000 epochs. For RIO we use the default setup in the reference code provided. Finally, for KNIFE, we employ 256 modes and train for 1000 epochs, cross-validating the learning rate and weight decay. All algorithms except RIO utilize the Adam optimizer~\cite{DBLP:journals/corr/KingmaB14} and batch normalization~\cite{pmlr-v37-ioffe15}. Additional details can be found in Appendix \ref{sec:details}. With the conditional estimators we implement the baselines following the procedure outlined in Appendix \ref{app:calc_probs}.
For the diversity detectors DV-Y and DV-D we use a neural network with four hidden layers of 64 neurons. We train for 25 epochs and choose the optimal learning rate using only a validation set of 20\% of the data. The best parameters are chosen to maximize the AUROC of the validation set. The symmetry in the output is achieved by averaging two replicas of the output of the network with exchanged inputs.
\vspace{-5mm}
\paragraph{Considered metrics}
We measure the performance of the detectors using the mean and standard deviation over $10$ seeds of the  metrics: \textbf{False Positive Rate} (FPR, fraction of samples classified as $\eps$-bad that are $\eps$-good) at 90\% of  \textbf{True Positive Rate} (TPR, fraction of samples classified as $\eps$-bad but are $\eps$-bad), labeled as FPR at TPR 90\% (lower is better);    \textbf{Area under the Receiver Operating Characteristic curve (AUROC)} which relates  the FPR to the TPR (perfect detector = 100\%)~\cite{AUROC_2004}.

\paragraph{Connection to Conformal Estimates} In general, conformal approaches cannot be directly applied to solve the issue we address. Specifically, the guarantees provided by conformal predictions, in terms of marginal coverage, do not offer any insight into the conditional probability required for constructing a detector. Nevertheless, to provide more insight, 
we discuss the main differences and propose a detection approach  based on conformalized quantile regression~\cite{Romano2019}. This analysis is relegated to \cref{app:conformal} due to space limitations.


\subsection{Discussion}
In Tables \ref{table:auroc_abs} and \ref{table:auroc_rel}, we present the AUROC (mean and standard deviation, normalized to 100) for the absolute and relative dispersion metrics, respectively. Tables \ref{table:fpr_abs} and \ref{table:fpr_rel} showcase the FPR at 90\% (mean and standard deviation, where lower values are better). Due to the large number of considered algorithms (6 baselines and 6 competing algorithms), we highlight the baselines that perform best for at least one value of $\eps$ and do the same for the proposed algorithms. This way, for each $\eps$, we compare the best baseline with the best among the proposed algorithms. The complete tables are available in Appendix \ref{app:tables}. In the tables with the AUROC values (Tables \ref{table:auroc_abs} and \ref{table:auroc_rel}) we also include the mean and standard deviation of the percentage of samples in the test set that are $\eps$-bad for each value of $\eps$ as reference.  Overall, it is observed that, in most cases, using the \textbf{DV algorithms attains better performance in terms of AUROC}, specifically up to 3.4/100 and 3.7/100 for the absolute and relative error metrics, respectively, compared with the best baselines. On the other hand,  DV algorithms perform, at worst, comparable or only marginally worse than competing algorithms. 
For some datasets, the \textbf{DV algorithms achieve a reduction of up to 50\% in FPR at TPR 90\%}, while in the worst case the performance is comparable. 

The ultimate measure of a detector's performance is determined by
the conditional distribution of $\mathbf{Y}|\mathbf{\Xvec}$. 
Particularly for small datasets, the fairness in
data splits plays a crucial role. 
Therefore, in Tables 1-4 we present our results by ablating across different values of the tolerated error
epsilon. This approach mirrors the actions of a practitioner who must adjust the tolerance level to suit a
specific task and dataset. The key observation from the tables provided is that the proposed method
(DVIC+SQR…) consistently outperforms the considered baseline (SQR, …) across various metrics. 


The algorithms computed using the distribution of 
$Y|\xvec$ (B1-* and DV-Y-*) tend to outperform their counterparts 
using the distribution of $D|\xvec$, 
possibly due to how the regressor is used during training in the 
second case. However, there are cases where the algorithms 
based on the estimates of $D|\Xvec=\xvec$ perform better. 
A detailed examination of the complete tables 
in Appendix \ref{app:tables} reveals significant variations 
in baseline performance across different methods 
for each dataset, likely due to the alignment of 
underlying assumptions of each algorithm with respect to 
the inherent distributions in the data. 
The parameter $\gamma$ plays a pivotal role, serving as a parameter
that governs the balance between TPR and FPR, i.e., the trade-off between accurately identifying positives
and minimizing false alarms. Thus, the choice of this 
parameter should be tailored to the
requirements of each specific application. 
To select $\gamma$, we recommend utilizing a calibration set,
consisting of representative data to guide this selection process.
A similar rationale applies to the choice of $\epsilon$, 
which denotes
the tolerance for error. Generally, 
the selection of $\epsilon$ depends on the task at hand, 
and practitioners
should choose it based on the level of error they deem 
acceptable for their specific use-case.

The proposed framework is inherently versatile, applicable 
to any type of regressor. This flexibility is one of its 
primary advantages. As demonstrated in our experimental 
results, its overall performance is generally comparable to,
 and often significantly better than, the baseline. However, 
 the extent of performance improvement depends on the specific 
 data distribution inherent in the task under consideration. 
 A distinct advantage of our detector is its data-driven approach, 
 which accounts for the data distribution.

 Bayesian analysis and related methods, such as the mixture of experts' paradigms, provide uncertainty estimates that can inform our proposed solution. Utilizing these estimates, like error distribution estimates, is of great interest and warrants future research. We plan to investigate the performance of Bayesian methods as a baseline and integrate them with our regression error detection method. Specifically, we aim to explore how the uncertainty estimates generated by Bayesian techniques can complement our approach, potentially enhancing the accuracy and robustness of our detection framework.

Finally, we add an analysis concerning
the computational time requirements of the algorithms, and the specific of the
hardware we utilized in the experiments
in Appendix \ref{app:computationTimes}.

\section{Summary and Concluding Remarks}
We proposed a novel framework to analyze and detect potential anomalies in the operation of regression algorithms. We introduced the notion of unreliability in regression, i.e., when the output of the regressor exceeds a specified discrepancy. We estimate the discrepancy density and measure its statistical diversity using our proposed data-driven score. Our framework demonstrates its effectiveness in compensating for estimation inaccuracies of the density of a target variable given the input, showcasing its potential for improving the reliability of the regressor's predictions in diverse applications, across various datasets and error metrics, which emphasizes its robustness and versatility.
Many existing rejection-based algorithms require simultaneous training with the regressor, which differs from our framework. As a future direction, it would be interesting to explore different setups and evaluation frameworks to compare them with our proposed solution.
Furthermore, extensions of our framework to other types of data 
(e.g. medical images, biometric, and multidimensional data), as well as other types of networks is a promising avenue for future research.

 \section*{Acknowledgements}
 
 This work has been supported by the project PSPC AIDA: 2019-PSPC-09 funded by BPI-France. This work was granted access to the HPC/AI resources of IDRIS under the allocation 2022 - AD011012803R1 made by GENCI.

\section*{Impact Statement}

This paper presents work whose goal is to advance the field of Machine Learning. There are many potential societal consequences of our work, none of which we feel must be specifically highlighted here.



\bibliography{main}
\bibliographystyle{bst}

\newpage
\appendix
\onecolumn

\section{Analysis of the  Example and Error Distributions of Real Datasets} \label{app:example}

In our example we consider a regressor whose difference random variable ${Y - f_{\Dset_n}(x) | X=x}$  is distributed as $\mathcal{N}(b(x), \sigma(x))$. Thus, defining the error variable $E = |Y-f_{\Dset_n}(x)|$ we may compute the probability that the error exceed the predefined threshold as:
\begin{equation}
\Prob\left(E > \eps |X=x\right) = 1- \left[ \Phi\left(\frac{\eps-b(x)}{\sigma(x)}\right)-\Phi\left(\frac{-\eps- b(x)}{\sigma(x)}\right) \right] ,\label{eq:Pg_E}
\end{equation}
where $\Phi$ is the cumulative distribution function of a standard normal random variable. From there we can compute the density of the error distribution from the difference variable $Y-f_{\Dset_n}(x)$ as:
\begin{equation}
    p_{E|X} (e|x) = p_{Y-f_{\Dset_n}(x)}(e) + p_{Y-f_{\Dset_n}(x)}(-e), \label{eq:density_E}
\end{equation}
where, as mentioned before, ${Y - f_{\Dset_n}(x) | X=x} \sim \mathcal{N}(b(x), \sigma(x))$.
For each possible $x\in\Xset $, comparing the probability in \cref{eq:Pg_E} with the detection threshold $\gamma$, we can identify the critical region of points where the regressor is deemed unreliable by the optimal detector.

\begin{figure}[h!]\centering
\includegraphics[width=0.5\columnwidth,trim=.5cm 0 0.45cm 0 ,clip ]{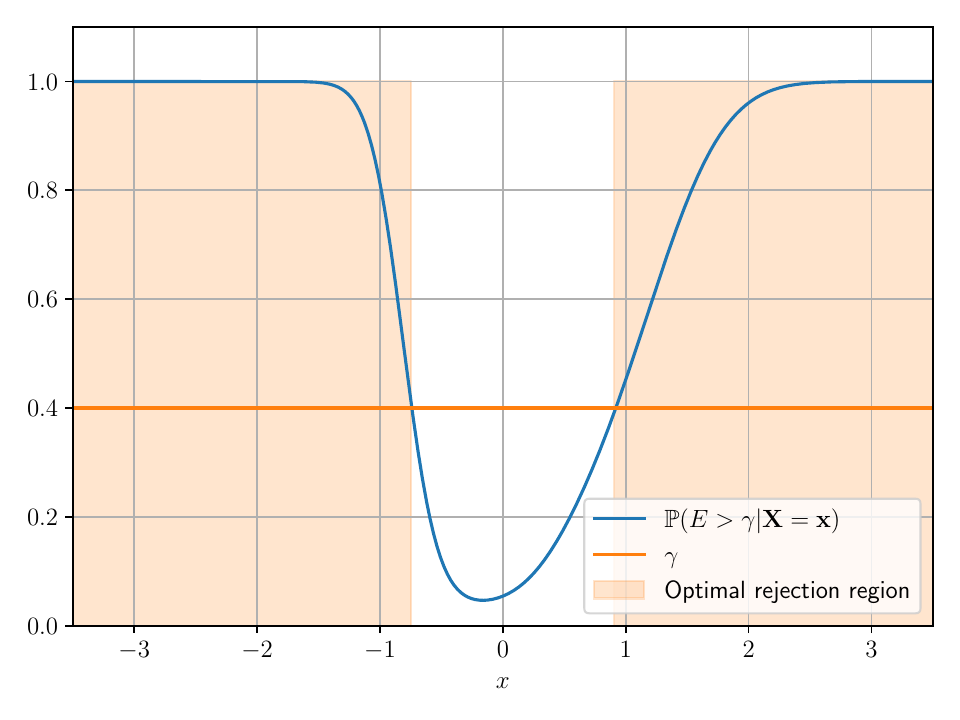} \caption{Optimal rejection region for the example in Section \ref{sec:example}.} \label{fig:opt_reg}
\end{figure}

For the specific example in Fig. \ref{fig:patt_toy}, we have taken $\sigma(x) = 0.05 (1+(x+0.2)^2)$ and $b(x) = 0.1(x-0.2)^3$ and chosen $\gamma=0.4$. With this we can compute  the rejection region which consists on the input points $x$ where the regressor is deemed unreliable (see Fig. \ref{fig:opt_reg}).
To compute the plots in Figs. \ref{fig:patt_toy}(a) and (b), we randomly choose 50 points for $X\sim \mathcal{N}(0,1)$, classify them as good and bad and compute the theoretical density \cref{eq:density_E}. We then compute the error product density \cref{eq:pE1E2} for each point in each class and average them within the class. 

To find the estimates in Figs. \ref{fig:patt_toy} (c) and (d), we generate a dataset $\{(x_i, e_i)\}_{i=1}^{1000}$ of input data and error samples. We then train a simple neural network architecture to learn the conditional quantile function of $E|X=x$ from this dataset. This network is used as a generative model to obtain an histogram of the conditional error distribution for the 50 input points used to generate Fig. \ref{fig:patt_toy}(a) and (b). We then estimate the error product distributions \cref{eq:pE1E2} and compute the  average of the estimates of the  joint error distributions.

\begin{figure}[t!]\centering
\begin{subfigure}[]{0.35\linewidth}
\includegraphics[width=1\columnwidth,trim=0cm 0 0.0cm 0 ,clip ]{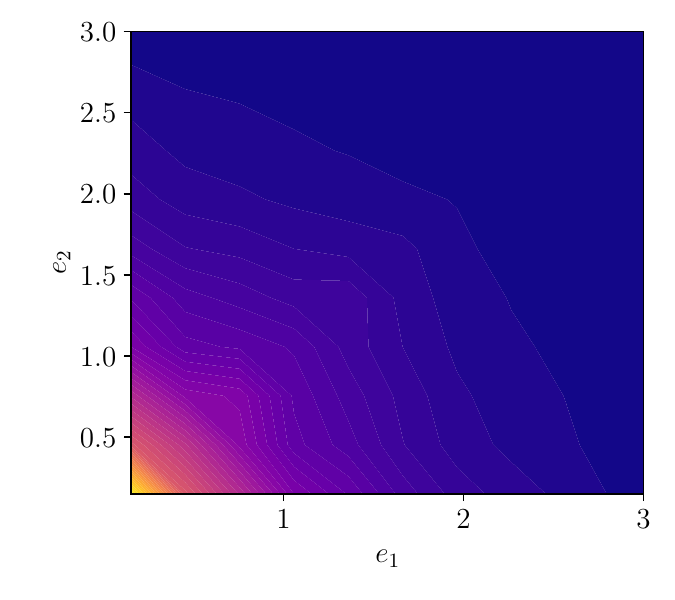}
\subcaption{Naval dataset. Average joint error distribution for points that are $\eps$-good in the test set.  $\eps=0.1$. Estimated using KNIFE.}
\end{subfigure}\hspace{0.3cm}
\begin{subfigure}[]{0.35\linewidth}
\raisebox{0.cm}{\includegraphics[width=1\columnwidth,trim=0cm 0cm 0.0cm 0,clip ]{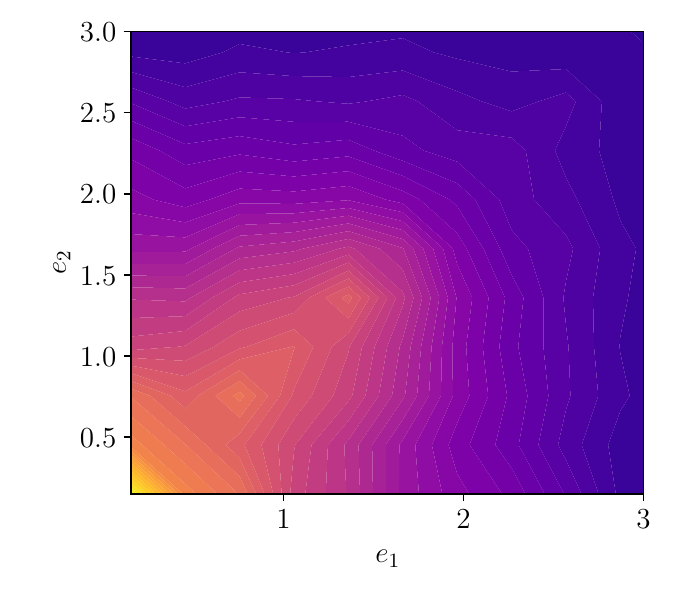}}
\subcaption{Naval dataset. Average joint error distribution for points that are $\eps$-bad in the test set. $\eps=0.1$. Estimated using KNIFE.}
\end{subfigure}
\vfill
\begin{subfigure}[t!]{0.35\linewidth}
\includegraphics[width=1\columnwidth,trim=0cm 0 0cm 0 ,clip ]{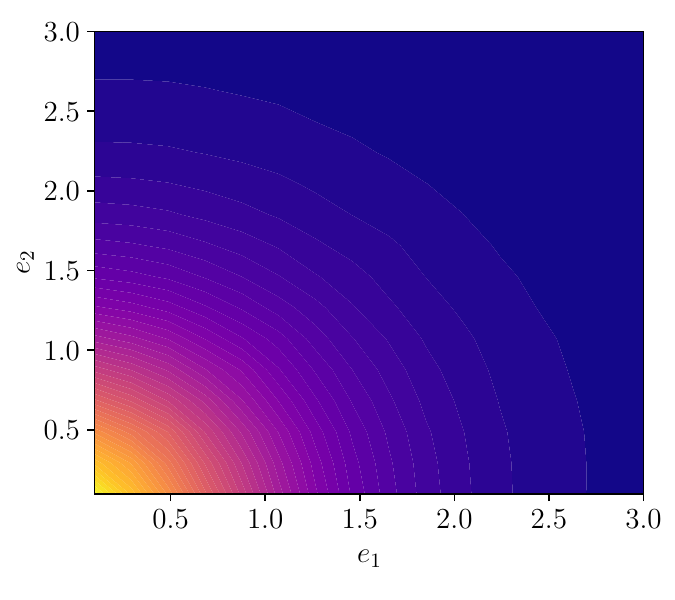}
\subcaption{Kin8nm dataset. Average joint error distribution for points that are $\eps$-good in the test set. $\eps=0.8$. Estimated using CG.}
\end{subfigure}\hspace{0.3cm}
\begin{subfigure}[t!]{0.35\linewidth}
\includegraphics[width=1\columnwidth,trim=.0cm 0 0cm 0,clip ]{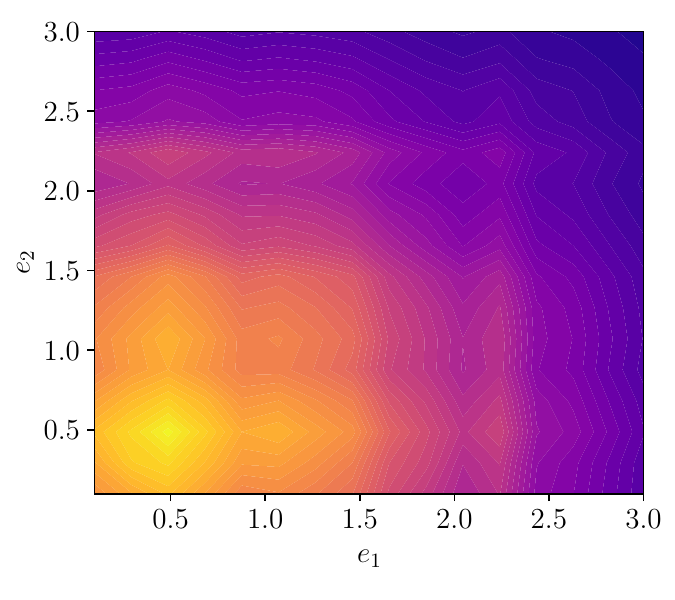}
\subcaption{Kin8nm dataset. Average joint error distribution for points that are $\eps$-bad in the test set. $\eps=0.8$. Estimated using CG.}
\end{subfigure}
\caption{Examples of average joint error distributions given by \cref{eq:pE1E2} for the absolute error discrepancy function $|Y-f_{\Dset_n}(x)|$ using the regressors trained in Section \ref{sec:numerical}. The patterns are obtained from the corresponding test set and the distributions of the discrepancy variables $D(Y, f_{\Dset_n}(x))$ are used. 
} \label{fig:pats_abs}
\end{figure}

We can follow a similar approach with real datasets, although establishing a ground truth is not possible in this scenario. In Fig. \ref{fig:pats_abs}, we employ the regressors and distribution estimation algorithms (specifically, SQR, CG, KNIFE) trained in Section \ref{sec:numerical} to estimate the conditional error distribution, focusing on the absolute error metric $|Y-f_{\Dset_n}(\xvec)|$. The distribution estimators are trained on the training set of the regressor, and subsequently, the error product distributions \cref{eq:pE1E2} are estimated on its test set. As observed in the simple example of Section \ref{sec:example}, the classes of good and bad points exhibit distinct behaviors in terms of the error product distributions.

\begin{figure}[t!]\centering
\begin{subfigure}[]{0.35\linewidth}
\includegraphics[width=1.03\columnwidth,trim=0cm 0 0.0cm 0 ,clip ]{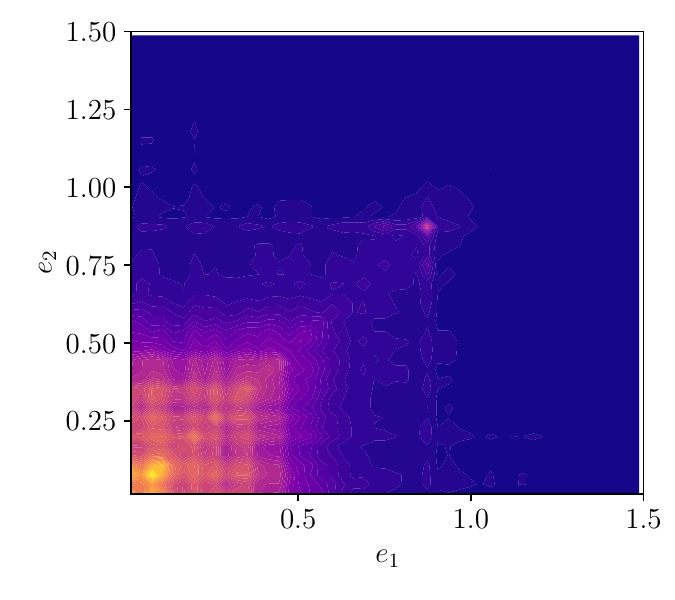}
\subcaption{Boston dataset. Average joint error distribution for points that are $\eps$-good in the test set.  $\eps=0.25$. Estimated using SQR.}
\end{subfigure}\hspace{0.3cm}
\begin{subfigure}[]{0.35\linewidth}
\raisebox{0.cm}{\includegraphics[width=1.03\columnwidth,trim=0cm 0cm 0.0cm 0,clip ]{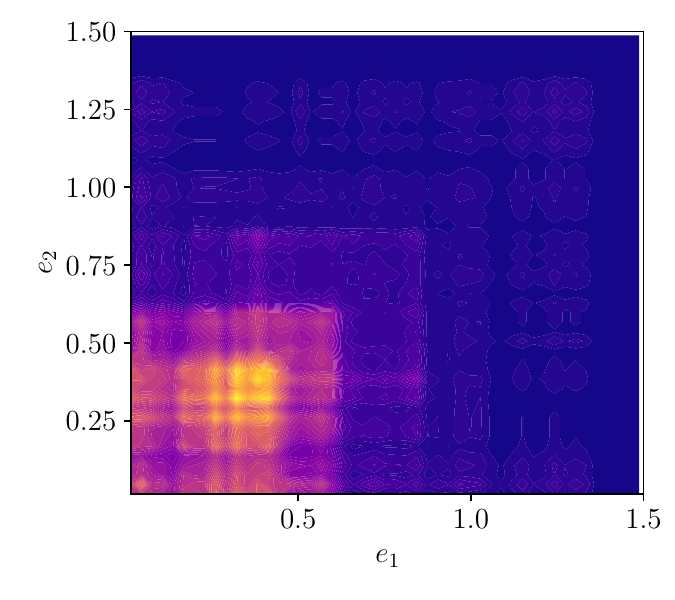}}
\subcaption{Boston dataset. Average joint error distribution for points that are $\eps$-bad in the test set. $\eps=0.25$. Estimated using SQR.}
\end{subfigure}
\vfill
\begin{subfigure}[t!]{0.35\linewidth}
\includegraphics[width=1.0\columnwidth,trim=0cm 0 0cm 0 ,clip ]{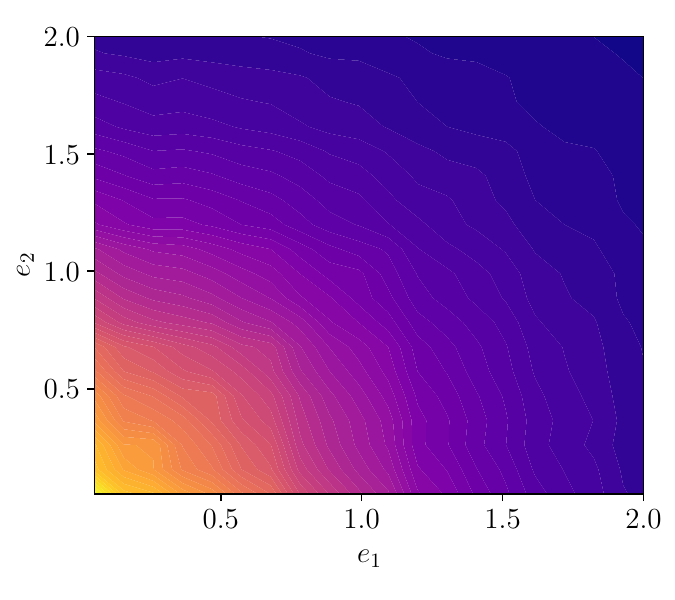}
\subcaption{Energy dataset. Average joint error distribution for points that are $\eps$-good in the test set. $\eps=0.05$. Estimated using CG.}
\end{subfigure}\hspace{0.3cm}
\begin{subfigure}[t!]{0.35\linewidth}
\includegraphics[width=1\columnwidth,trim=.0cm 0 0cm 0,clip ]{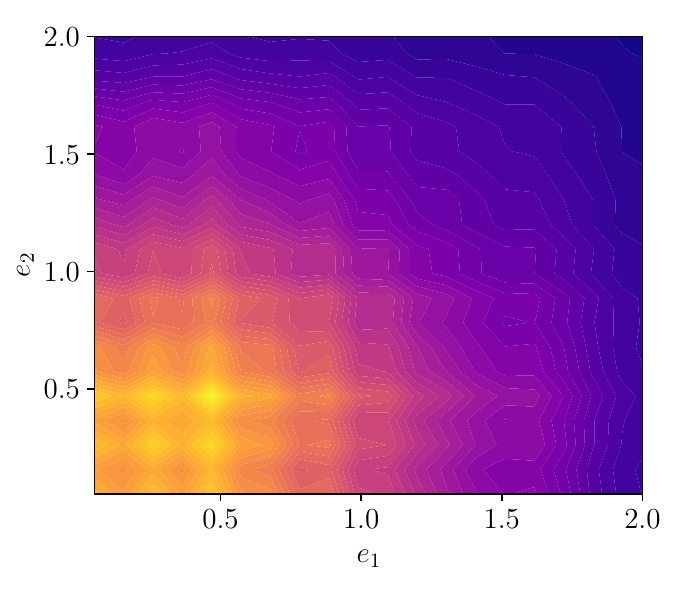}
\subcaption{Energy dataset. Average joint error distribution for points that are $\eps$-bad in the test set. $\eps=0.05$. Estimated using CG.}
\end{subfigure}
\caption{Examples of average joint error distributions given by \cref{eq:pE1E2} for the relative error discrepancy function $|Y-f_{\Dset_n}|/|f_{\Dset_n}|$ using the regressors trained in Section \ref{sec:numerical}. The patterns are obtained from the corresponding test set and the distributions of the discrepancy variables $D(Y, f_{\Dset_n})$ are used. \label{fig:pats_rel}
}
\end{figure}

In Fig. \ref{fig:pats_rel}, a similar procedure is applied to two datasets, namely \emph{energy} and \emph{Boston}, for the relative error metric $|Y-f_{\Dset_n}(\xvec)|/|f_{\Dset_n}(\xvec)|$ described in Section \ref{sec:numerical}. Once again, different patterns for each class can be observed in this case.



\section{Proofs of Propositions}
\subsection{Proof of \cref{prop:most_powerful_discriminator}}
\label{sec:proof-most-powerful}

Define the random variable $J \triangleq A_{\eps}(\Xvec,\Yvec)$ using \cref{eq:Aeps}. The null hypothesis of $\eps$-goodness is then simply $J = 0$ and the alternative is $J = 1$. We wish to apply the fundamental lemma of Neyman and Pearson~(Thm.~3.2.1 in~\cite{Lehmann2005Testing}) and choose the measure $\mu \triangleq P_{\Xvec} \times \delta_{\{0,1\}}$ on $\Xset \times \{0,1\}$, where $P_{\Xvec}$ is the distribution of $\Xvec$ and $\delta_{\{0,1\}}$ is the counting measure.

For $\Prob(J = j)> 0$, we can write the density $p_{\Xvec}(\xvec|J=j)$ of $\Xvec$ given $J=j$ w.r.t.\ $P_{\Xvec}$ as
\begin{align}
  p_{\Xvec}(\xvec|J=j) &= \frac{1}{\Prob(J = j)} \frac{d \Prob(\Xvec, J)}{d\mu}(\xvec, j) 
                       = \frac{1}{\Prob(J = j)} \Prob(J = j|\Xvec = \xvec). \label{eq:density}
\end{align}
The null hypothesis is rejected if and only if $\Prob(J = 1|\Xvec = \xvec) = P_B(\xvec ; \eps) > \gamma$.
This condition can now be rewritten as $p_{\Xvec}(\xvec|J=1) > k \ p_{\Xvec}(\xvec|J=0)$ using \cref{eq:density}, where 
$k
= \frac{\gamma P\{J = 0\}}{(1-\gamma) P\{J = 1\}}$.
However, this is exactly the sufficient condition for the most powerful test by Thm.~3.2.1 in~\cite{Lehmann2005Testing}.

\subsection{Proof of Proposition \ref{prop:Hopt}}\label{sec:proof_div}
We can consider the diversity for the regressor obtained from the symmetric function:
\begin{equation}
    h_p(u, v)  \triangleq \bigone\{u > \eps\} \bigone\{v > \eps\}. \label{eq:hopt}
\end{equation}
Then, since in the definition of $   \mathbb{H}(\xvec)$ in Eq. \cref{eq:divC} $\Yvec_1$ and $\Yvec_2$ are drawn independently according to $\Yvec|\Xvec=\xvec$, we have:
\begin{align}
       \mathbb{H}(\xvec) &= \Ex\left[ \bigone\{d(\Yvec_1, f_{\Dset_n}(\xvec)) > \eps\} \times \bigone
    \{d(\Yvec_2, f_{\Dset_n}(\xvec)) > \eps\} |\Xvec=\xvec \right]\nonumber \\
    &= \Ex\left[ \bigone\{d(\Yvec_1, f_{\Dset_n}(\xvec)) > \eps\}|\Xvec = \xvec\right] \times
    \Ex\left[\bigone\{d(\Yvec_2, f_{\Dset_n}(\xvec)) > \eps\} |\Xvec=\xvec \right]\nonumber \\
    &= P^2_B(\xvec; \eps).
\end{align}
Since $P^2_B(\xvec;\eps)$ is an increasing function of $P_B(\xvec ; \eps)$, this detector is equivalent to the optimal one.

\newpage
\section{Algorithms based on the discrepancy variable} \label{sec:algorithms}
In this Appendix, we present the baseline Algorithm \ref{alg:baseline2} which  is similar to Algorithm \ref{alg:baseline1}, but based on the estimation of the conditional distribution of $D|\Xvec =\xvec$. We also present Algorithm \ref{alg:div_discD} (DV-D), which is similar to Algorithm \ref{alg:div_discY} (DV-Y) but also based on the estimation of the conditional distribution of $D|\Xvec=\xvec$.
\begin{algorithm}[h!]
   \caption{Baseline based on the estimation of the conditional distribution of $D(\Yvec, \Xvec)|\Xvec=\xvec$}
   \label{alg:baseline2}
\begin{algorithmic}
   \STATE {\bfseries Input:} a trained regressor $f_{\Dset_n}$, trained on the dataset $\Dset_n = \{(\xvec_i, y_i)\}_{i=1}^n$. A detection threshold $\gamma \in [0, \ 1]$.
   
\vspace{1mm}
\STATE \textbf{Training:}
\STATE \textbf{\hspace{3mm} Step 1:} Using $\Dset_n$ define a new discrepancy samples dataset:
\vspace{-3mm}
$$\tilde{\Dset}_{n} \triangleq \{ (\xvec_i, d(\yvec_i, f_{\Dset_n}(\xvec_{i}))\}_{i=1}^n.$$
\STATE \textbf{\hspace{3mm} Step 2:} Use  $\tilde{\Dset}_{n}$ to estimate  the conditional distribution of $D(\Yvec, \Xvec)|\Xvec$.
   \vspace{1mm}
\STATE \textbf{After training:} Given an input test sample $\xvec$:
\STATE \textbf{\hspace{3mm} Step 1:} Estimate $\hat{P}_B(\xvec;\eps)$ using \cref{eq:Pg2}
\STATE \textbf{\hspace{3mm} Step 2:} Decide that the regressor is $\eps$-bad at $\xvec$ if $\hat{P}_B(\xvec;\eps) > \gamma.$
\end{algorithmic}
\end{algorithm}

\begin{algorithm}[h!]
   \caption{\textbf{-- DV-Y-D}: Diversity discriminator based on the estimation of the conditional distribution of  $D(\Yvec, \Xvec)|\Xvec=\xvec$}
   \label{alg:div_discD}
\begin{algorithmic} 
   \STATE {\bfseries Input:} A regressor $f_{\Dset_n}$ trained on  dataset $\Dset_n $, dissimilarity threshold $\gamma$.
   
\STATE \textbf{Training:}
\STATE \textbf{\hspace{3mm}Step 1:} Using $\Dset_n$ define a new discrepancy  dataset:
\vspace{-2mm}
$$\tilde{\Dset}_{n} \triangleq \{ (\xvec_i, d(\yvec_i, f_{\Dset_n}(\xvec_{i}))\}_{i=1}^n.$$
\STATE \textbf{\hspace{3mm} Step 2:} Use  $\tilde{\Dset}_{n}$ to obtain an estimate of the distribution of the discrepancy variable $D(\Yvec, \Xvec)|\Xvec$.
\STATE \textbf{\hspace{3mm}Step 3:} separate good and bad samples in $\Dset_n$:
\vspace{-2mm}
\begin{align*}
    \mathcal{G} \triangleq \{ (\xvec, \yvec) \in \Dset_n : d(\yvec, f_{\Dset_n}(\xvec)) \leq \eps\},\\
    \mathcal{B} \triangleq \{ (\xvec,\yvec) \in \Dset_n : d(\yvec, f_{\Dset_n}(\xvec)) > \eps\}.
\end{align*}

\STATE \textbf{\hspace{3mm}Step 4:} learn the function $h$ that defined the diversity metric $\mathbb{H}$ for the discriminator:
\FOR {$i$ in $1,..,N_\text{epochs}$} 
\FOR {each $(\xvec, d(\yvec, f_{\Dset_n}(\xvec)))\in \tilde{\Dset}_n$}
\STATE Generate two extended datasets of $N_u$ samples:
\vspace{-1mm}
\begin{alignat*}{2}
    \tilde{\Dset}_1(\xvec) = \{\bar{d}_1,\dots,\bar{d}_{N_u}\} \hspace{2mm} & \hspace{2mm}\tilde{\Dset}_2(\xvec) = \{\tilde{d}_1,\dots ,\tilde{d}_{N_u}\}
\end{alignat*}
of discrepancy samples, where each $\bar{d}_i, \tilde{d}_i$ is generated i.i.d.\ with  $\hat{F}_{\mathbf{D}|\Xvec=\xvec}$ from Step 2.
\STATE Estimate the diversity metric for $\xvec$:
$$   \mathbb{H}(\xvec) = \frac{1}{N_u} \sum_{i=1}^{N_u} h(\tilde{d}_i,\bar{d}_i),$$\vspace{-4mm}
\ENDFOR
\STATE Update $h$ to minimize:
$$\mathcal{L} = \frac{1}{2} \frac{1}{|\mathcal{G}|}\sum_{\xvec \in \mathcal{G}}    \mathbb{H}(\xvec) - \frac{1}{2} \frac{1}{|\mathcal{B}|} \sum_{\xvec \in \mathcal{B}}    \mathbb{H}(\xvec)$$
such that $h$ is symmetric and $h:\mathbb{R}^2_{\geq 0}  \rightarrow [0, 1]$.
\ENDFOR 

\STATE \textbf{After training:} Given an input test sample $\xvec$:
\STATE \textbf{\hspace{3mm} Step 1:} compute $   \mathbb{H}(\xvec)$ via a Monte Carlo experiment like in Step 4 of training. 
\STATE \textbf{\hspace{3mm} Step 2:} Decide that the regressor will be $\eps$-bad at $\xvec$ if   $\mathbb{H}(\xvec) > \gamma$.
\end{algorithmic}
\end{algorithm}

\section{Additional Simulation Details} \label{sec:details}
\subsection{Hyperparameters}
For the CG, SQR and KNIFE the following setups were used:
\begin{itemize}
    \item CG: we averaged 10 models and performed 5-fold cross-validation over the learning rate in $\{0.01, 0.001, 0.0001\}$. Each model was trained for 150 epochs. For the mean and variance a neural network with a single hidden layer of 64 neurons was used. 

    \item SQR: we used a neural network with 3 hidden layers of 64 neurons. We performed 5-fold cross-validation over the learning rate in $\{10^{-3}, 5\times 10^{-4}, 10^{-4}, 5\times 10^{-5}, 10^{-5}\}$ and weight decay in $\{0, 0.025, 0.05, 0.075, 0.1\}$. The models were trained for 1000 epochs.
    
    \item KNIFE: we used 256 modes for the Gaussian mixture, and performed 5-fold cross-validation over the learning rate in $\{0.001, 0.0005, 0.0001\}$ and weight decay in $\{0, 0.0125, 0.025\}$. The models were trained for 1000 epochs. 
\end{itemize}
For the implementation of the DV-Y and DV-D algorithms, a neural network with 4 hidden layers, each containing 64 neurons, was employed. The network's symmetry was achieved by averaging the output of two copies of the network fed with interchanged inputs. In all Monte Carlo experiments to compute $\mathbb{H}$, 20000 uniform random variables were used. For learning the $h$ function in DV-Y and DV-D algorithms, the learning rates were chosen based on a 20\% validation set, within the following ranges:
\begin{itemize}
\item \emph{yacht}, \emph{energy}: $\{5 \times 10^{-3}, 10^{-3}, 5\times 10^{-4}\}$;
\item \emph{concrete}, \emph{kin8nm}, \emph{naval}: $\{10^{-4}, 5\times 10^{-5}, 10^{-5}\}$;
\item \emph{power}, \emph{wine}: $\{10^{-3}, 5\times10^{-4}, 10^{-4}\}$;
\item \emph{Boston}: $\{5\times 10^{-4}, 5\times10^{-5}, 10^{-5}\}$.
\end{itemize}

\subsection{Computations of $\hat{P}_B(\xvec;\eps)$ for the baseline algorithms}
\label{app:calc_probs}

For the computations of the baselines in Section \ref{sec:numerical} where the output variable $\Yvec$ is scalar, $\Yset \subseteq \Real$, we computed $\hat{P}_B(\xvec;\eps)$ in the simulations using the following expressions:

\begin{itemize}
    \item For the absolute error and a scalar output $Y$ we have:
    \begin{align}
	P_B(\xvec ; \eps) &= \int_\Yset \bigone \{d_a(y,f_{\Dset_n}(\xvec)) \ge \eps\} \ dF_{Y|\Xvec=\xvec}(y) \nonumber\\
 &= 1 - \Prob \left( f_{\Dset_n}(\xvec) -\eps \leq Y\leq f_{\Dset_n}(\xvec) + \eps | \Xvec= \xvec\right) \nonumber\\
	&= 1 - F_{Y|\Xvec=\xvec}(f_{\Dset_n}(\xvec)+\eps) + F_{Y|\Xvec=\xvec}(f_{\Dset_n}(\xvec)-\eps). \label{eq:Pga_1}
\end{align}
\item For the relative error and a scalar output $Y$ similar computations show that:
\begin{equation}
P_B(\xvec ; \eps) = 1 - F_{Y|\Xvec=\xvec}(f_{\Dset_n}(\xvec) + \eps | f_{\Dset_n}(\xvec)|) + F_{Y|\Xvec=\xvec}(f_{\Dset_n}(\xvec) - \eps | f_{\Dset_n}(\xvec)|).
\end{equation}
\end{itemize}
In the case of CG and KNIFE, which are parametric models involving Gaussian random variables, we can compute these probabilities in closed form. For SQR, the algorithm provides the quantile function (inverse cumulative distribution function). To compute the probability, we then need to numerically invert this function to find the required probabilities, which is straightforward to do since the quantile function is monotonous and defined on $ [0, 1] $.

\section{Analysis of computational cost}
\label{app:computationTimes}
The computational cost of training the DV detectors 
can be separated in two
components:

\begin{enumerate}
  \item In the first step, an algorithm that predicts the conditional distribution either of $\mathbf{Y}|\mathbf{X}$
  (for DV-Y) or of the discrepancy random variables (for DV-D) has to be considered. The cost associated with
  this stage comes from the algorithm of choice. For instance, the cost of SQR, KNIFE and the conditional
  Gaussian method are different, and they depend on the complexity of the chosen model for each case. This
  training has to be performed in an equal manner for both the baseline algorithms and the algorithm with
  the diversity coefficients.
  \item In the second step, the diversity coefficient algorithm needs to be trained. This process involves training a
  model that learns the function $h$, which is essentially a mapping from $\mathbb{R} \times \mathbb{R} \to
  \mathbb{R}_{\geq 0}$. In other words, $h$ operates as a system with two real inputs and one output. For
  our experiments, we employed a neural network with 4 hidden layers, each containing 64 neurons. To obtain
  the Monte Carlo estimation of the diversity coefficients $\mathbb{H}$ given by (12) in the paper, we utilize
  the model trained in the first step as a generative model. This model is used to generate a large number of
  realizations of the discrepancy random variables, typically 20,000 in our experiments. Subsequently, these
  realizations serve as inputs to the neural network responsible for generating the samples of $h$. Since the
  network that learns $h$ only has two inputs, the computational complexity of this training process does not
  surpass that of training a network with inputs equivalent to a square image of approximately 141 pixels per
  side, with the added step of generating the scalar samples.
\end{enumerate}
In Table \ref{tab:timetable} we show some examples of training time for the diversity coefficients in different scenarios.
\input{tables2/compute_times.txt}
\FloatBarrier

\section{Connection to conformal prediction}
\label{app:conformal}

While there are some common 
elements with conformal prediction, the
objective of this work differs. In contrast, conformal approaches cannot be 
directly applied to solve the problem we tackle,
although we acknowledge and verify the useful properties of 
marginal coverage they possess. The key differences are 
as follows:
\begin{enumerate}
  \item Our goal is to detect unreliable regressor behavior as 
        measured by an application dependent error margin
        ($\epsilon$) and a discrepancy metric. For our baselines 
        we use a probabilistic model, such as quantile
        regression (SQR) or others (conditional gaussian, KNIFE), 
        to estimate the conditional probability of the
        unreliable behavior event:
        $$P
        _
        B(\mathbf{x}) = \mathbb{P}(d(\mathbf{Y}, f
        _{\mathcal{D}_
        n} (\mathbf{x})) > \epsilon | \mathbf{X} =
        \mathbf{x}).$$
  \item Using this probability, we approximate the optimal detector.
        Conformal approaches provide a prediction interval 
        that contains the desired variables with certain marginal
        guarantees. To achieve this, a desired value for the 
        marginal coverage probability (denoted as $1-\alpha$) is
        selected, and then an interval (similar in spirit to the 
        interval defined by $\epsilon$) that satisfies this
        marginal coverage condition is obtained. In other words, 
        a value of $\epsilon$ is determined such that:
        $$ \mathbb{P}(d(\mathbf{X}, f
        _{\mathcal{D}_
        n}(\mathbf{X})) \leq \epsilon) \geq 1-\alpha. $$
\end{enumerate}

Thus, in conformal learning the marginal probability is 
fixed and the width $\epsilon$ is found, to guarantee
marginal coverage.
The analysis above demonstrates that the problem we are tackling 
differs from the problem addressed by
conformal predictions. Moreover, the guarantees provided by 
conformal predictions, in terms of marginal
coverage, do not offer any insight into the conditional 
probability required for constructing our baselines.


Nevertheless, we explore the use of conformal approaches within
our framework,  by constructing the following approach: given value 
of $\epsilon$ and an input $\mathbf{x}$:
\begin{enumerate}
  \item[a.] Construct a conformal prediction such that the interval 
        matches the condition of being $\epsilon$-good, i.e.
        {$d\left(\mathbf{Y}, f_{\mathcal{D}_n}(\mathbf{x})\right) \leq \epsilon$}.
  \item[b.] Obtain the marginal coverage probability for this event, 
        denoted as $P_c(\mathbf{x})$.
  \item[c.] Use $1-P_c(\mathbf{x})$ as a proxy 
        for the conditional probability $P_B(\mathbf{x})$.
\end{enumerate}

This approach does not have any theoretical guarantees 
since we approximate the conditional probability via a
marginal coverage probability of an interval of the chosen 
width provided by a specific conformal algorithm.
To test this algorithm we need a conformal algorithm which 
satisfies two conditions:
\begin{enumerate}
  \item The conformal interval has to be possibly non-symmetric 
  (allow for different tails) ;
  \item The interval has to depend on $\mathbf{x}$. The fact that not all conformal techniques satisfy this\cite{Vovk2005}, shows in fact that this construction is not inherent to these approaches.
\end{enumerate}

In this case, we rely on the well-known method introduced in 
\cite{Romano2019}, specifically Theorem 2, which provides an
algorithm that satisfies the desired properties and allows 
us to construct the conformal intervals based on any
distribution of the coverage of its tails.
This algorithm starts from a preconstructed confidence interval 
and then applies a conformal correction. This
means that for each algorithm used in our baseline 
(i.e., SQR, KNIFE, conditional Gaussian) we can construct a
“conformal” version, which consists of computing the baseline with 
the procedure described in the previous
paragraph. We compare the performance of these algorithms with the 
ones already present in the paper, which
are indicated with the suffix CF (e.g., B1-SQR-CF denotes Baseline 
algorithm 1, with quantiles computed with SQR
and the conformal correction). The results can be seen in the 
tables below. In particular, for the tables in this section, we
report the original results on the left and the conformal
correction results on the right.
As we see in the results, both for the relative error and absolute 
error metric we performed 144 experiments
with the 8 datasets, 3 epsilons and the 6 baselines. 
In terms of AUROC, the conformal modification of the
baseline only results in improvement in 52\% of the experiments. 
This shows that the effects of this type of
correction are uncertain since none of the theoretical 
guarantees afforded by conformal intervals apply.
\begin{table}[ht!] \scriptsize \centering  \caption{AUROC for the  relative error discrepancy metric   for  the small  datasets. B1, B2 correspond to baseline Algorithms. DV-Y and DV-D  correspond to the proposed algorithms. CF corresponds to the baseline with the conformal correction.}\vskip 0.15in
	
\begin{minipage}[t!]{0.45\linewidth}
\input{tablesCF/small_rel_AUROC_base.txt}
\end{minipage}
\begin{minipage}[t!]{0.45\linewidth}
\input{tablesCF/small_rel_AUROC_CF.txt}
\end{minipage}
\end{table}

\begin{table}[ht!] \centering \scriptsize \caption{FPR  at TPR 90\% for the  relative error discrepancy metric   for  the small  datasets. B1, B2 correspond to baseline Algorithms. DV-Y and DV-D  correspond to the proposed algorithms. CF corresponds to the baseline with the conformal correction.}\vskip 0.15in
\begin{minipage}[t!]{0.45\linewidth}
\input{tablesCF/small_rel_FPR_base.txt}
\end{minipage}
\begin{minipage}[t!]{0.45\linewidth}
\input{tablesCF/small_rel_FPR_CF.txt}
\end{minipage}

\end{table}

\newpage

\begin{table}[ht!] \scriptsize \centering  \caption{AUROC for the  absolute error discrepancy metric   for  the small  datasets. B1, B2 correspond to baseline Algorithms. DV-Y and DV-D  correspond to the proposed Algorithms. CF corresponds to the baseline with the conformal correction.}
	\vskip 0.15in
\begin{minipage}[t!]{0.45\linewidth}
	\input{tablesCF/small_abs_AUROC_base.txt}

\end{minipage}
\begin{minipage}[t!]{0.45\linewidth}
	\input{tablesCF/small_abs_AUROC_CF.txt}

\end{minipage}

\end{table}

\begin{table}[ht!] \centering \scriptsize \caption{FPR  at TPR 90\% for the  absolute error discrepancy metric   for  the small  datasets. B1, B2 correspond to baseline Algorithms. DV-Y and DV-D  correspond to the proposed Algorithms. CF corresponds to the baseline with the conformal correction.}
	\vskip 0.15in
\begin{minipage}[t!]{0.45\linewidth}
	\input{tablesCF/small_abs_FPR_base.txt}
\end{minipage}	
\begin{minipage}[t!]{0.45\linewidth}

\input{tablesCF/small_abs_FPR_CF.txt}
\end{minipage}	
\end{table}

\newpage
\begin{table}[ht!] \scriptsize \centering  \caption{AUROC for the  relative error discrepancy metric   for  the large  datasets. B1, B2 correspond to baseline Algorithms. DV-Y and DV-D  correspond to the proposed Algorithms. CF corresponds to the baseline with the conformal correction.}\vskip 0.15in
\begin{minipage}[t!]{0.45\linewidth}
	\input{tablesCF/large_rel_AUROC_base.txt}
\end{minipage}	
\begin{minipage}[t!]{0.45\linewidth}
	\input{tablesCF/large_rel_AUROC_CF.txt}
\end{minipage}	
\end{table}

\newpage
\begin{table}[ht!] \centering \scriptsize \caption{FPR  at TPR 90\% for the  relative error discrepancy metric   for  the large  datasets. B1, B2 correspond to baseline Algorithms. DV-Y and DV-D  correspond to the proposed Algorithms. CF corresponds to the baseline with the conformal correction.}\vskip 0.15in
\begin{minipage}[t!]{0.45\linewidth}
	\input{tablesCF/large_rel_FPR_base.txt}
\end{minipage}	
\begin{minipage}[t!]{0.45\linewidth}
	\input{tablesCF/large_rel_FPR_CF.txt}
\end{minipage}	
\end{table}

\newpage

\begin{table}[ht!] \scriptsize \centering  \caption{AUROC for the  absolute error discrepancy metric   for  the large  datasets. B1, B2 correspond to baseline Algorithms. DV-Y and DV-D  correspond to the proposed Algorithms.CF corresponds to the baseline with the conformal correction.}
	\vskip 0.15in
\begin{minipage}[t!]{0.45\linewidth}
	\input{tablesCF/large_abs_AUROC_base.txt}
\end{minipage}	
\begin{minipage}[t!]{0.45\linewidth}
	\input{tablesCF/large_abs_AUROC_CF.txt}
\end{minipage}		
\end{table}
	
\newpage
\begin{table}[ht!] \centering \scriptsize \caption{FPR  at TPR 90\% for the  absolute error discrepancy metric   for  the large  datasets. B1, B2 correspond to baseline Algorithms. DV-Y and DV-D  correspond to the proposed Algorithms. CF corresponds to the baseline with the conformal correction.}
	\vskip 0.15in
\begin{minipage}[t!]{0.45\linewidth}
	\input{tablesCF/large_abs_FPR_base.txt}
\end{minipage}	
\begin{minipage}[t!]{0.45\linewidth}
	\input{tablesCF/large_abs_FPR_CF.txt}
\end{minipage}	

\end{table}

\newpage
\FloatBarrier
\section{Complete Tables of Results}
\label{app:completeTables}
This Appendix contains the full tables of simulations results over all the baseline algorithms and the proposed approaches. There are several interesting observations that can be made from the tables:
\begin{itemize}
    \item Looking at the baselines we see that there is not a clear winner in terms of the distribution estimation algorithms over the datasets. This is probably because the underlying assumptions of each method may be better adjusted to model specific datasets. In most cases the baselines based on the estimation of the distribution of $\Yvec|\Xvec=\xvec$ perform better than the counterparts based on $D|\Xvec=\xvec$ but in some cases this is not the case.
    \item The DV algorithms in general perform better than their counterpart based on the same conditional estimators. In some cases achieving gains of up to 18.5/100 in average AUROC. This means that DV is capable of substantially improving the detection results in many cases and reduce the variability across dataset splits. 

\end{itemize}

\label{app:tables}

\input{tables2/small_abs.txt}
\input{tables2/large_abs.txt}

\input{tables2/small_rel.txt}
\input{tables2/large_rel.txt}

\end{document}

%% file: tables2/all_abs_red.txt
\begin{table}[t!] \small \centering  \caption{AUROC (mean and standard deviation, normalized to 100) for the  absolute error discrepancy metric. B1, B2 correspond to baseline Algorithms \ref{alg:baseline1} and \ref{alg:baseline2} resp. DV-Y and DV-D  correspond to the proposed Algorithms \ref{alg:div_discY} and \ref{alg:div_discD} resp. For each dataset we report the baselines that are the best for at least one $\eps$, and the same for the proposed algorithms. The full tables are in Appendix \ref{app:tables}. } \label{table:auroc_abs}

\vskip 0.1in
\begin{tabular}{|c|c|ccc|} \hline\multirow{5}{*}{\rotatebox[origin=c]{90}{\footnotesize \textbf{naval}}}
& \cellcolor{tbl1} $\bm{\eps}$ & \cellcolor{tbl1} \textbf{0.175}  & \cellcolor{tbl1} \textbf{0.2}  & \cellcolor{tbl1} \textbf{0.225}  \\ \hhline{~----}
& \textbf{Avg. \% bad} \cellcolor{tbl1} & \cellcolor{tbl1}2.6 [0.9] & \cellcolor{tbl1}1.4 [0.7] & \cellcolor{tbl1}0.9 [0.4] \\ \hhline{~----}
& B2-SQR& 93.9 [3.8]& 93.9 [4.2]& 88.8 [7.2]\\ 
& B2-KNIFE& 91.6 [3.5]& 93.4 [3.6]& 95.2 [2.7]\\ 
& DV-D-SQR \cellcolor{tbl2}& \textbf{96.7} [1.3]\cellcolor{tbl2}& \textbf{97.3} [1.8]\cellcolor{tbl2}& \textbf{96.2} [3.6]\cellcolor{tbl2}\\ \hline
\multirow{6}{*}{\rotatebox[origin=c]{90}{\footnotesize \textbf{power}}}
& \cellcolor{tbl1} $\bm{\eps}$ & \cellcolor{tbl1} \textbf{0.3}  & \cellcolor{tbl1} \textbf{0.35}  & \cellcolor{tbl1} \textbf{0.4}  \\ \hhline{~----}
& \textbf{Avg. \% bad} \cellcolor{tbl1} & \cellcolor{tbl1}14.1 [1.0] & \cellcolor{tbl1}9.2 [1.0] & \cellcolor{tbl1}6.4 [0.7] \\ \hhline{~----}
& B1-SQR& 70.3 [2.5]& 71.4 [2.5]& 71.1 [1.6]\\ 
& B2-SQR& 70.4 [2.6]& 70.9 [2.7]& 71.1 [3.8]\\ 
& DV-Y-SQR \cellcolor{tbl2}& 70.8 [2.7]\cellcolor{tbl2}& \textbf{72.2} [2.5]\cellcolor{tbl2}& \textbf{72.8} [3.3]\cellcolor{tbl2}\\ 
& DV-D-SQR \cellcolor{tbl2}& \textbf{71.5} [2.5]\cellcolor{tbl2}& \textbf{72.2} [2.7]\cellcolor{tbl2}& 72.0 [3.4]\cellcolor{tbl2}\\ \hline
\multirow{4}{*}{\rotatebox[origin=c]{90}{\footnotesize \textbf{kin8nm}}}
& \cellcolor{tbl1} $\bm{\eps}$ & \cellcolor{tbl1} \textbf{0.6}  & \cellcolor{tbl1} \textbf{0.7}  & \cellcolor{tbl1} \textbf{0.8}  \\ \hhline{~----}
& \textbf{Avg. \% bad} \cellcolor{tbl1} & \cellcolor{tbl1}4.2 [0.6] & \cellcolor{tbl1}2.2 [0.5] & \cellcolor{tbl1}1.1 [0.3] \\ \hhline{~----}
& B1-CG& \textbf{78.6} [2.7]& \textbf{81.9} [3.7]& \textbf{86.9} [5.4]\\ 
& DV-Y-CG \cellcolor{tbl2}& \textbf{78.6} [2.7]\cellcolor{tbl2}& 81.7 [3.8]\cellcolor{tbl2}& 86.7 [5.4]\cellcolor{tbl2}\\ \hline
\multirow{6}{*}{\rotatebox[origin=c]{90}{\footnotesize \textbf{wine}}}
& \cellcolor{tbl1} $\bm{\eps}$ & \cellcolor{tbl1} \textbf{0.1}  & \cellcolor{tbl1} \textbf{0.3}  & \cellcolor{tbl1} \textbf{0.5}  \\ \hhline{~----}
& \textbf{Avg. \% bad} \cellcolor{tbl1} & \cellcolor{tbl1}87.4 [4.5] & \cellcolor{tbl1}65.2 [6.1] & \cellcolor{tbl1}46.8 [5.0] \\ \hhline{~----}
& B1-SQR& 91.9 [5.1]& 84.2 [5.2]& 71.3 [3.6]\\ 
& B1-KNIFE& 94.5 [2.5]& 86.0 [4.7]& 70.3 [4.9]\\ 
& DV-Y-SQR \cellcolor{tbl2}& 94.3 [4.3]\cellcolor{tbl2}& 86.1 [4.3]\cellcolor{tbl2}& \textbf{72.2} [3.6]\cellcolor{tbl2}\\ 
& DV-Y-KNIFE \cellcolor{tbl2}& \textbf{95.2} [2.1]\cellcolor{tbl2}& \textbf{87.3} [3.8]\cellcolor{tbl2}& \textbf{72.2} [2.9]\cellcolor{tbl2}\\ \hline
\multirow{5}{*}{\rotatebox[origin=c]{90}{\footnotesize \textbf{concrete}}}
& \cellcolor{tbl1} $\bm{\eps}$ & \cellcolor{tbl1} \textbf{0.2}  & \cellcolor{tbl1} \textbf{0.5}  & \cellcolor{tbl1} \textbf{0.6}  \\ \hhline{~----}
& \textbf{Avg. \% bad} \cellcolor{tbl1} & \cellcolor{tbl1}36.7 [5.6] & \cellcolor{tbl1}9.6 [2.8] & \cellcolor{tbl1}6.4 [2.0] \\ \hhline{~----}
& B1-SQR& 73.1 [3.8]& 81.1 [8.3]& 76.3 [14.2]\\ 
& B2-CG& 69.8 [7.1]& 78.7 [9.6]& 81.5 [9.1]\\ 
& DV-Y-SQR \cellcolor{tbl2}& \textbf{74.4} [3.4]\cellcolor{tbl2}& \textbf{83.5} [8.3]\cellcolor{tbl2}& \textbf{82.1} [8.4]\cellcolor{tbl2}\\ \hline
\multirow{4}{*}{\rotatebox[origin=c]{90}{\footnotesize \textbf{energy}}}
& \cellcolor{tbl1} $\bm{\eps}$ & \cellcolor{tbl1} \textbf{0.025}  & \cellcolor{tbl1} \textbf{0.0625}  & \cellcolor{tbl1} \textbf{0.1}  \\ \hhline{~----}
& \textbf{Avg. \% bad} \cellcolor{tbl1} & \cellcolor{tbl1}61.2 [5.0] & \cellcolor{tbl1}25.8 [5.2] & \cellcolor{tbl1}9.5 [3.4] \\ \hhline{~----}
& B1-CG& 72.5 [6.8]& 81.2 [5.5]& 81.7 [5.4]\\ 
& DV-Y-CG \cellcolor{tbl2}& \textbf{75.1} [6.0]\cellcolor{tbl2}& \textbf{82.0} [6.9]\cellcolor{tbl2}& \textbf{82.5} [7.7]\cellcolor{tbl2}\\ \hline
\multirow{4}{*}{\rotatebox[origin=c]{90}{\footnotesize \textbf{boston}}}
& \cellcolor{tbl1} $\bm{\eps}$ & \cellcolor{tbl1} \textbf{0.25}  & \cellcolor{tbl1} \textbf{0.3}  & \cellcolor{tbl1} \textbf{0.35}  \\ \hhline{~----}
& \textbf{Avg. \% bad} \cellcolor{tbl1} & \cellcolor{tbl1}31.8 [2.7] & \cellcolor{tbl1}24.1 [5.8] & \cellcolor{tbl1}21.2 [5.4] \\ \hhline{~----}
& B1-SQR& 72.8 [7.2]& 77.1 [6.0]& 76.8 [5.6]\\ 
& DV-Y-SQR \cellcolor{tbl2}& \textbf{73.7} [7.7]\cellcolor{tbl2}& \textbf{77.6} [6.7]\cellcolor{tbl2}& \textbf{78.6} [5.4]\cellcolor{tbl2}\\ \hline
\multirow{4}{*}{\rotatebox[origin=c]{90}{\footnotesize \textbf{yacht}}}
& \cellcolor{tbl1} $\bm{\eps}$ & \cellcolor{tbl1} \textbf{0.025}  & \cellcolor{tbl1} \textbf{0.05}  & \cellcolor{tbl1} \textbf{0.075}  \\ \hhline{~----}
& \textbf{Avg. \% bad} \cellcolor{tbl1} & \cellcolor{tbl1}62.6 [9.6] & \cellcolor{tbl1}35.8 [11.3] & \cellcolor{tbl1}25.5 [10.6] \\ \hhline{~----}
& B1-CG& 92.4 [4.4]& 94.1 [5.7]& 96.3 [3.0]\\ 
& DV-Y-CG \cellcolor{tbl2}& \textbf{94.3} [3.3]\cellcolor{tbl2}& \textbf{94.4} [5.7]\cellcolor{tbl2}& \textbf{96.6} [2.9]\cellcolor{tbl2}\\ \hline
\end{tabular}\end{table}

\begin{table}[t!]  \centering \small \caption{FPR  at TPR 90\%  (mean and standard deviation, lower is better) for the  absolute error discrepancy metric.  B1, B2 correspond to baseline Algorithms \ref{alg:baseline1} and \ref{alg:baseline2} resp. DV-Y and DV-D  correspond to our Algorithms \ref{alg:div_discY} and \ref{alg:div_discD} resp. For each dataset we report the baselines that are the best for at least one $\eps$, and the same for the proposed algorithms. The full tables are in Appendix \ref{app:tables}.}\label{table:fpr_abs}
\vskip 0.1in
\begin{tabular}{|c|c|ccc|} \hline\multirow{5}{*}{\rotatebox[origin=c]{90}{\footnotesize \textbf{naval}}}
& \cellcolor{tbl1} $\bm{\eps}$ & \cellcolor{tbl1} \textbf{0.175}  & \cellcolor{tbl1} \textbf{0.2}  & \cellcolor{tbl1} \textbf{0.225}  \\ \hhline{~----}
& B2-KNIFE& 0.22 [0.11]& 0.19 [0.13]& \textbf{0.12} [0.06]\\ 
& B2-CG& 0.20 [0.04]& 0.19 [0.05]& 0.15 [0.07]\\ 
& DV-D-SQR \cellcolor{tbl2}& \textbf{0.11} [0.05]\cellcolor{tbl2}& \textbf{0.09} [0.05]\cellcolor{tbl2}& 0.15 [0.18]\cellcolor{tbl2}\\ 
& DV-D-KNIFE \cellcolor{tbl2}& 0.22 [0.08]\cellcolor{tbl2}& 0.19 [0.10]\cellcolor{tbl2}& \textbf{0.12} [0.10]\cellcolor{tbl2}\\ \hline
\multirow{4}{*}{\rotatebox[origin=c]{90}{\footnotesize \textbf{power}}}
& \cellcolor{tbl1} $\bm{\eps}$ & \cellcolor{tbl1} \textbf{0.3}  & \cellcolor{tbl1} \textbf{0.35}  & \cellcolor{tbl1} \textbf{0.4}  \\ \hhline{~----}
& B1-SQR& 0.69 [0.06]& 0.69 [0.08]& 0.74 [0.11]\\ 
& DV-Y-SQR \cellcolor{tbl2}& \textbf{0.68} [0.05]\cellcolor{tbl2}& \textbf{0.67} [0.07]\cellcolor{tbl2}& \textbf{0.67} [0.09]\cellcolor{tbl2}\\ 
& DV-D-SQR \cellcolor{tbl2}& \textbf{0.68} [0.05]\cellcolor{tbl2}& 0.71 [0.06]\cellcolor{tbl2}& 0.75 [0.07]\cellcolor{tbl2}\\ \hline
\multirow{3}{*}{\rotatebox[origin=c]{90}{\footnotesize \textbf{kin8nm}}}
& \cellcolor{tbl1} $\bm{\eps}$ & \cellcolor{tbl1} \textbf{0.6}  & \cellcolor{tbl1} \textbf{0.7}  & \cellcolor{tbl1} \textbf{0.8}  \\ \hhline{~----}
& B1-CG& \textbf{0.48} [0.09]& \textbf{0.44} [0.12]& \textbf{0.33} [0.20]\\ 
& DV-Y-CG \cellcolor{tbl2}& \textbf{0.48} [0.10]\cellcolor{tbl2}& \textbf{0.44} [0.12]\cellcolor{tbl2}& \textbf{0.33} [0.21]\cellcolor{tbl2}\\ \hline
\multirow{4}{*}{\rotatebox[origin=c]{90}{\footnotesize \textbf{wine}}}
& \cellcolor{tbl1} $\bm{\eps}$ & \cellcolor{tbl1} \textbf{0.1}  & \cellcolor{tbl1} \textbf{0.3}  & \cellcolor{tbl1} \textbf{0.5}  \\ \hhline{~----}
& B1-KNIFE& 0.14 [0.14]& 0.49 [0.14]& \textbf{0.73} [0.08]\\ 
& DV-Y-SQR \cellcolor{tbl2}& \textbf{0.11} [0.13]\cellcolor{tbl2}& 0.51 [0.15]\cellcolor{tbl2}& \textbf{0.72} [0.06]\cellcolor{tbl2}\\ 
& DV-Y-KNIFE \cellcolor{tbl2}& 0.13 [0.12]\cellcolor{tbl2}& \textbf{0.46} [0.15]\cellcolor{tbl2}& \textbf{0.72} [0.07]\cellcolor{tbl2}\\ \hline
\multirow{6}{*}{\rotatebox[origin=c]{90}{\footnotesize \textbf{concrete}}}
& \cellcolor{tbl1} $\bm{\eps}$ & \cellcolor{tbl1} \textbf{0.2}  & \cellcolor{tbl1} \textbf{0.5}  & \cellcolor{tbl1} \textbf{0.6}  \\ \hhline{~----}
& B1-SQR& \textbf{0.67} [0.10]& 0.77 [0.39]& 0.90 [0.30]\\ 
& B2-KNIFE& 0.73 [0.13]& 0.54 [0.18]& \textbf{0.45} [0.27]\\ 
& B1-CG& 0.69 [0.10]& \textbf{0.44} [0.13]& 0.49 [0.18]\\ 
& DV-Y-SQR \cellcolor{tbl2}& \textbf{0.67} [0.10]\cellcolor{tbl2}& \textbf{0.43} [0.21]\cellcolor{tbl2}& 0.51 [0.22]\cellcolor{tbl2}\\ 
& DV-D-CG \cellcolor{tbl2}& \textbf{0.67} [0.12]\cellcolor{tbl2}& 0.51 [0.22]\cellcolor{tbl2}& 0.49 [0.23]\cellcolor{tbl2}\\ \hline
\multirow{3}{*}{\rotatebox[origin=c]{90}{\footnotesize \textbf{energy}}}
& \cellcolor{tbl1} $\bm{\eps}$ & \cellcolor{tbl1} \textbf{0.025}  & \cellcolor{tbl1} \textbf{0.0625}  & \cellcolor{tbl1} \textbf{0.1}  \\ \hhline{~----}
& B1-CG& 0.61 [0.11]& \textbf{0.42} [0.14]& \textbf{0.39} [0.14]\\ 
& DV-Y-CG \cellcolor{tbl2}& \textbf{0.59} [0.10]\cellcolor{tbl2}& \textbf{0.42} [0.13]\cellcolor{tbl2}& \textbf{0.39} [0.18]\cellcolor{tbl2}\\ \hline
\multirow{4}{*}{\rotatebox[origin=c]{90}{\footnotesize \textbf{boston}}}
& \cellcolor{tbl1} $\bm{\eps}$ & \cellcolor{tbl1} \textbf{0.25}  & \cellcolor{tbl1} \textbf{0.3}  & \cellcolor{tbl1} \textbf{0.35}  \\ \hhline{~----}
& B1-SQR& 0.69 [0.17]& 0.60 [0.17]& 0.64 [0.25]\\ 
& B1-KNIFE& 0.70 [0.14]& 0.64 [0.16]& \textbf{0.57} [0.12]\\ 
& DV-Y-SQR \cellcolor{tbl2}& \textbf{0.64} [0.21]\cellcolor{tbl2}& \textbf{0.53} [0.20]\cellcolor{tbl2}& \textbf{0.56} [0.24]\cellcolor{tbl2}\\ \hline
\multirow{3}{*}{\rotatebox[origin=c]{90}{\footnotesize \textbf{yacht}}}
& \cellcolor{tbl1} $\bm{\eps}$ & \cellcolor{tbl1} \textbf{0.025}  & \cellcolor{tbl1} \textbf{0.05}  & \cellcolor{tbl1} \textbf{0.075}  \\ \hhline{~----}
& B1-CG& \textbf{0.16} [0.12]& \textbf{0.11} [0.12]& 0.16 [0.30]\\ 
& DV-Y-CG \cellcolor{tbl2}& \textbf{0.16} [0.10]\cellcolor{tbl2}& \textbf{0.11} [0.12]\cellcolor{tbl2}& \textbf{0.10} [0.07]\cellcolor{tbl2}\\ \hline
\end{tabular}\end{table}

%% file: tables2/all_rel_red.txt
\begin{table}[t!]  \small \centering  \caption{AUROC (mean and standard deviation, normalized to 100) for the  relative error discrepancy metric. B1, B2 correspond to baseline Algorithms \ref{alg:baseline1} and \ref{alg:baseline2} resp. DV-Y and DV-D  correspond to the proposed Algorithms \ref{alg:div_discY} and \ref{alg:div_discD} resp. For each dataset we report the baselines that are the best for at least one $\eps$, and the same for the proposed algorithms. The full tables are in Appendix \ref{app:tables}.}\label{table:auroc_rel}
\vskip 0.1in
\begin{tabular}{|c|c|ccc|} \hline\multirow{5}{*}{\rotatebox[origin=c]{90}{\footnotesize \textbf{naval}}}
& \cellcolor{tbl1} $\bm{\eps}$ \textbf{[\%]} & \cellcolor{tbl1} \textbf{0.15}\%  & \cellcolor{tbl1} \textbf{0.175}\%  & \cellcolor{tbl1} \textbf{0.2}\%  \\ \hhline{~----}
& \textbf{Avg. \% bad} \cellcolor{tbl1} & \cellcolor{tbl1}1.5 [0.6] & \cellcolor{tbl1}0.8 [0.4] & \cellcolor{tbl1}0.5 [0.3] \\ \hhline{~----}
& B2-SQR& 94.8 [4.1]& 87.6 [7.2]& 90.2 [9.7]\\ 
& B2-KNIFE& 92.4 [3.8]& 95.1 [3.2]& 96.7 [2.7]\\ 
& DV-D-SQR \cellcolor{tbl2}& \textbf{97.3} [1.8]\cellcolor{tbl2}& \textbf{97.0} [2.3]\cellcolor{tbl2}& \textbf{98.0} [1.8]\cellcolor{tbl2}\\ \hline
\multirow{5}{*}{\rotatebox[origin=c]{90}{\footnotesize \textbf{power}}}
& \cellcolor{tbl1} $\bm{\eps}$ \textbf{[\%]} & \cellcolor{tbl1} \textbf{1}\%  & \cellcolor{tbl1} \textbf{1.25}\%  & \cellcolor{tbl1} \textbf{1.5}\%  \\ \hhline{~----}
& \textbf{Avg. \% bad} \cellcolor{tbl1} & \cellcolor{tbl1}18.2 [0.8] & \cellcolor{tbl1}10.9 [0.9] & \cellcolor{tbl1}6.6 [0.8] \\ \hhline{~----}
& B2-SQR& 71.4 [2.9]& 72.3 [2.9]& 71.6 [2.6]\\ 
& DV-Y-SQR \cellcolor{tbl2}& 70.4 [2.3]\cellcolor{tbl2}& 72.5 [2.9]\cellcolor{tbl2}& \textbf{72.6} [3.2]\cellcolor{tbl2}\\ 
& DV-D-SQR \cellcolor{tbl2}& \textbf{71.6} [2.7]\cellcolor{tbl2}& \textbf{72.8} [2.5]\cellcolor{tbl2}& 72.3 [2.3]\cellcolor{tbl2}\\ \hline
\multirow{4}{*}{\rotatebox[origin=c]{90}{\footnotesize \textbf{kin8nm}}}
& \cellcolor{tbl1} $\bm{\eps}$ \textbf{[\%]} & \cellcolor{tbl1} \textbf{10}\%  & \cellcolor{tbl1} \textbf{20}\%  & \cellcolor{tbl1} \textbf{30}\%  \\ \hhline{~----}
& \textbf{Avg. \% bad} \cellcolor{tbl1} & \cellcolor{tbl1}34.4 [1.3] & \cellcolor{tbl1}12.2 [1.2] & \cellcolor{tbl1}5.2 [0.6] \\ \hhline{~----}
& B1-CG& 76.8 [1.8]& \textbf{85.8} [1.7]& \textbf{91.2} [1.9]\\ 
& DV-Y-CG \cellcolor{tbl2}& \textbf{77.0} [1.9]\cellcolor{tbl2}& \textbf{85.8} [1.8]\cellcolor{tbl2}& \textbf{91.2} [2.0]\cellcolor{tbl2}\\ \hline
\multirow{6}{*}{\rotatebox[origin=c]{90}{\footnotesize \textbf{wine}}}
& \cellcolor{tbl1} $\bm{\eps}$ \textbf{[\%]} & \cellcolor{tbl1} \textbf{5}\%  & \cellcolor{tbl1} \textbf{10}\%  & \cellcolor{tbl1} \textbf{15}\%  \\ \hhline{~----}
& \textbf{Avg. \% bad} \cellcolor{tbl1} & \cellcolor{tbl1}62.6 [4.8] & \cellcolor{tbl1}32.8 [3.6] & \cellcolor{tbl1}17.0 [2.1] \\ \hhline{~----}
& B1-SQR& 82.1 [5.2]& \textbf{61.8} [3.8]& 65.1 [6.2]\\ 
& B1-KNIFE& 82.7 [5.9]& 60.6 [6.3]& 69.9 [6.0]\\ 
& DV-Y-SQR \cellcolor{tbl2}& 84.2 [4.8]\cellcolor{tbl2}& 59.6 [3.4]\cellcolor{tbl2}& 64.4 [8.4]\cellcolor{tbl2}\\ 
& DV-Y-KNIFE \cellcolor{tbl2}& \textbf{84.4} [5.0]\cellcolor{tbl2}& 59.3 [8.5]\cellcolor{tbl2}& \textbf{70.2} [7.0]\cellcolor{tbl2}\\ \hline
\multirow{4}{*}{\rotatebox[origin=c]{90}{\footnotesize \textbf{concrete}}}
& \cellcolor{tbl1} $\bm{\eps}$ \textbf{[\%]} & \cellcolor{tbl1} \textbf{10}\%  & \cellcolor{tbl1} \textbf{15}\%  & \cellcolor{tbl1} \textbf{20}\%  \\ \hhline{~----}
& \textbf{Avg. \% bad} \cellcolor{tbl1} & \cellcolor{tbl1}35.4 [6.2] & \cellcolor{tbl1}22.1 [5.1] & \cellcolor{tbl1}13.9 [3.4] \\ \hhline{~----}
& B1-SQR& 71.3 [5.3]& 74.8 [5.1]& 74.6 [6.3]\\ 
& DV-Y-SQR \cellcolor{tbl2}& \textbf{72.2} [5.4]\cellcolor{tbl2}& \textbf{75.3} [5.0]\cellcolor{tbl2}& \textbf{77.4} [4.6]\cellcolor{tbl2}\\ \hline
\multirow{7}{*}{\rotatebox[origin=c]{90}{\footnotesize \textbf{energy}}}
& \cellcolor{tbl1} $\bm{\eps}$ \textbf{[\%]} & \cellcolor{tbl1} \textbf{2}\%  & \cellcolor{tbl1} \textbf{3.5}\%  & \cellcolor{tbl1} \textbf{5}\%  \\ \hhline{~----}
& \textbf{Avg. \% bad} \cellcolor{tbl1} & \cellcolor{tbl1}46.8 [3.7] & \cellcolor{tbl1}21.0 [7.0] & \cellcolor{tbl1}10.5 [5.2] \\ \hhline{~----}
& B1-SQR& 71.1 [8.2]& 80.5 [6.4]& 91.3 [5.9]\\ 
& B1-CG& 74.9 [3.9]& 78.3 [6.8]& 81.9 [12.1]\\ 
& DV-Y-SQR \cellcolor{tbl2}& 73.1 [7.9]\cellcolor{tbl2}& 81.0 [5.9]\cellcolor{tbl2}& \textbf{91.5} [5.7]\cellcolor{tbl2}\\ 
& DV-Y-KNIFE \cellcolor{tbl2}& 73.5 [6.9]\cellcolor{tbl2}& \textbf{81.9} [5.2]\cellcolor{tbl2}& 87.4 [8.4]\cellcolor{tbl2}\\ 
& DV-Y-CG \cellcolor{tbl2}& \textbf{76.3} [3.2]\cellcolor{tbl2}& 80.1 [6.6]\cellcolor{tbl2}& 89.1 [7.9]\cellcolor{tbl2}\\ \hline
\multirow{4}{*}{\rotatebox[origin=c]{90}{\footnotesize \textbf{boston}}}
& \cellcolor{tbl1} $\bm{\eps}$ \textbf{[\%]} & \cellcolor{tbl1} \textbf{10}\%  & \cellcolor{tbl1} \textbf{15}\%  & \cellcolor{tbl1} \textbf{20}\%  \\ \hhline{~----}
& \textbf{Avg. \% bad} \cellcolor{tbl1} & \cellcolor{tbl1}33.7 [5.2] & \cellcolor{tbl1}20.8 [5.0] & \cellcolor{tbl1}14.5 [6.3] \\ \hhline{~----}
& B1-SQR& 72.2 [8.0]& 78.3 [7.5]& 81.9 [11.9]\\ 
& DV-Y-SQR \cellcolor{tbl2}& \textbf{74.0} [7.4]\cellcolor{tbl2}& \textbf{79.5} [8.4]\cellcolor{tbl2}& \textbf{85.6} [8.2]\cellcolor{tbl2}\\ \hline
\multirow{4}{*}{\rotatebox[origin=c]{90}{\footnotesize \textbf{yacht}}}
& \cellcolor{tbl1} $\bm{\eps}$ \textbf{[\%]} & \cellcolor{tbl1} \textbf{10}\%  & \cellcolor{tbl1} \textbf{15}\%  & \cellcolor{tbl1} \textbf{20}\%  \\ \hhline{~----}
& \textbf{Avg. \% bad} \cellcolor{tbl1} & \cellcolor{tbl1}59.4 [9.0] & \cellcolor{tbl1}46.1 [9.3] & \cellcolor{tbl1}39.0 [8.9] \\ \hhline{~----}
& B1-CG& \textbf{92.7} [4.4]& \textbf{95.7} [3.0]& \textbf{97.2} [2.1]\\ 
& DV-Y-CG \cellcolor{tbl2}& 92.1 [5.0]\cellcolor{tbl2}& 94.6 [3.5]\cellcolor{tbl2}& 96.2 [2.9]\cellcolor{tbl2}\\ \hline
\end{tabular} \vspace{-2mm}\end{table}

\begin{table}[t!]\centering \small \caption{FPR  at TPR 90\% (mean and standard deviation, lower is better) for the  relative error discrepancy metric. B1, B2 correspond to baseline Algorithms \ref{alg:baseline1} and \ref{alg:baseline2} resp. DV-Y and DV-D  correspond to the proposed Algorithms \ref{alg:div_discY} and \ref{alg:div_discD} resp. For each dataset we report the baselines that are the best for at least one $\eps$, and the same for the proposed algorithms. The full tables are in Appendix \ref{app:tables}.} \label{table:fpr_rel} \vskip 0.1in
\begin{tabular}{|c|c|ccc|} \hline\multirow{4}{*}{\rotatebox[origin=c]{90}{\footnotesize \textbf{naval}}}
& \cellcolor{tbl1} $\bm{\eps}$ \textbf{[\%]} & \cellcolor{tbl1} \textbf{0.15}\%  & \cellcolor{tbl1} \textbf{0.175}\%  & \cellcolor{tbl1} \textbf{0.2}\%  \\ \hhline{~----}
& B2-KNIFE& 0.22 [0.13]& 0.12 [0.07]& \textbf{0.09} [0.08]\\ 
& B2-CG& 0.16 [0.03]& 0.14 [0.07]& 0.11 [0.09]\\ 
& DV-D-SQR \cellcolor{tbl2}& \textbf{0.09} [0.08]\cellcolor{tbl2}& \textbf{0.08} [0.07]\cellcolor{tbl2}& \textbf{0.08} [0.09]\cellcolor{tbl2}\\ \hline
\multirow{4}{*}{\rotatebox[origin=c]{90}{\footnotesize \textbf{power}}}
& \cellcolor{tbl1} $\bm{\eps}$ \textbf{[\%]} & \cellcolor{tbl1} \textbf{1}\%  & \cellcolor{tbl1} \textbf{1.25}\%  & \cellcolor{tbl1} \textbf{1.5}\%  \\ \hhline{~----}
& B1-SQR& \textbf{0.68} [0.06]& \textbf{0.67}[0.07]& 0.77 [0.13]\\ 
& B1-CG& 0.82 [0.03]& 0.79 [0.05]& 0.76 [0.06]\\ 
& DV-Y-SQR \cellcolor{tbl2}& \textbf{0.67} [0.05]\cellcolor{tbl2}& \textbf{0.66} [0.05]\cellcolor{tbl2}& \textbf{0.67} [0.10]\cellcolor{tbl2}\\ \hline
\multirow{3}{*}{\rotatebox[origin=c]{90}{\footnotesize \textbf{kin8nm}}}
& \cellcolor{tbl1} $\bm{\eps}$ \textbf{[\%]} & \cellcolor{tbl1} \textbf{10}\%  & \cellcolor{tbl1} \textbf{20}\%  & \cellcolor{tbl1} \textbf{30}\%  \\ \hhline{~----}
& B1-CG& \textbf{0.57} [0.04]& \textbf{0.35} [0.04]& \textbf{0.19} [0.05]\\ 
& DV-Y-CG \cellcolor{tbl2}& \textbf{0.57} [0.04]\cellcolor{tbl2}& \textbf{0.35} [0.04]\cellcolor{tbl2}& \textbf{0.19} [0.05]\cellcolor{tbl2}\\ \hline
\multirow{3}{*}{\rotatebox[origin=c]{90}{\footnotesize \textbf{wine}}}
& \cellcolor{tbl1} $\bm{\eps}$ \textbf{[\%]} & \cellcolor{tbl1} \textbf{5}\%  & \cellcolor{tbl1} \textbf{10}\%  & \cellcolor{tbl1} \textbf{15}\%  \\ \hhline{~----}
& B1-KNIFE& 0.59 [0.13]& 0.79 [0.08]& \textbf{0.68} [0.13]\\ 
& DV-Y-KNIFE \cellcolor{tbl2}& \textbf{0.55} [0.15]\cellcolor{tbl2}& \textbf{0.76} [0.11]\cellcolor{tbl2}& \textbf{0.68} [0.15]\cellcolor{tbl2}\\ \hline
\multirow{4}{*}{\rotatebox[origin=c]{90}{\footnotesize \textbf{concrete}}}
& \cellcolor{tbl1} $\bm{\eps}$ \textbf{[\%]} & \cellcolor{tbl1} \textbf{10}\%  & \cellcolor{tbl1} \textbf{15}\%  & \cellcolor{tbl1} \textbf{20}\%  \\ \hhline{~----}
& B1-SQR& \textbf{0.63} [0.11]& 0.62 [0.21]& 0.76 [0.32]\\ 
& B2-CG& 0.73 [0.09]& 0.64 [0.17]& 0.62 [0.23]\\ 
& DV-Y-SQR \cellcolor{tbl2}& \textbf{0.63} [0.11]\cellcolor{tbl2}& \textbf{0.56} [0.09]\cellcolor{tbl2}& \textbf{0.47} [0.14]\cellcolor{tbl2}\\ \hline
\multirow{5}{*}{\rotatebox[origin=c]{90}{\footnotesize \textbf{energy}}}
& \cellcolor{tbl1} $\bm{\eps}$ \textbf{[\%]} & \cellcolor{tbl1} \textbf{2}\%  & \cellcolor{tbl1} \textbf{3.5}\%  & \cellcolor{tbl1} \textbf{5}\%  \\ \hhline{~----}
& B1-SQR& 0.76 [0.14]& 0.64 [0.20]& 0.35 [0.31]\\ 
& B1-CG& \textbf{0.62} [0.09]& \textbf{0.53} [0.19]& 0.40 [0.27]\\ 
& DV-Y-CG \cellcolor{tbl2}& \textbf{0.62} [0.11]\cellcolor{tbl2}& 0.54 [0.18]\cellcolor{tbl2}& \textbf{0.29} [0.25]\cellcolor{tbl2}\\ \hline
\multirow{4}{*}{\rotatebox[origin=c]{90}{\footnotesize \textbf{boston}}}
& \cellcolor{tbl1} $\bm{\eps}$ \textbf{[\%]} & \cellcolor{tbl1} \textbf{10}\%  & \cellcolor{tbl1} \textbf{15}\%  & \cellcolor{tbl1} \textbf{20}\%  \\ \hhline{~----}
& B1-KNIFE& 0.65 [0.11]& 0.61 [0.20]& 0.46 [0.30]\\ 
& B1-CG& 0.65 [0.22]& 0.56 [0.28]& 0.48 [0.33]\\ 
& DV-Y-SQR \cellcolor{tbl2}& \textbf{0.63} [0.18]\cellcolor{tbl2}& \textbf{0.52} [0.18]\cellcolor{tbl2}& \textbf{0.34} [0.15]\cellcolor{tbl2}\\ \hline
\multirow{4}{*}{\rotatebox[origin=c]{90}{\footnotesize \textbf{yacht}}}
& \cellcolor{tbl1} $\bm{\eps}$ \textbf{[\%]} & \cellcolor{tbl1} \textbf{10}\%  & \cellcolor{tbl1} \textbf{15}\%  & \cellcolor{tbl1} \textbf{20}\%  \\ \hhline{~----}
& B1-CG& \textbf{0.25} [0.21]& \textbf{0.17} [0.17]& \textbf{0.10} [0.12]\\ 
& DV-Y-SQR \cellcolor{tbl2}& 0.38 [0.25]\cellcolor{tbl2}& \textbf{0.17} [0.10]\cellcolor{tbl2}& 0.16 [0.10]\cellcolor{tbl2}\\ 
& DV-Y-CG \cellcolor{tbl2}& \textbf{0.26} [0.21]\cellcolor{tbl2}& 0.22 [0.21]\cellcolor{tbl2}& \textbf{0.10} [0.11]\cellcolor{tbl2}\\ \hline
\end{tabular} \vspace{-2mm}\end{table}

%% file: tables2/compute_times.txt
{\footnotesize
\begin{longtable}{|c|c|cccc|} 
\caption{Average training time for one epoch of training for each type of diversity coefficient using a Nvidia V100 GPU with 16GB of RAM. This time considers only the training of the diversity coefficients. The time differences are due to the different times required to generate the samples with each conditional model.} \label{tab:timetable}
\\
\hline
	\textbf{Dataset} & \textbf{Sample size} & \textbf{DV type} & \textbf{Conditional estimator} & \textbf{Discrepancy} & \textbf{Avg. time per it (sec.)} \\ \hline
	\multirow{12}{*}{yacht} & \multirow{12}{*}{308} & DV-Y &SQR &abs &1.0 \\ 
	& &  DV-Y &KNIFE &abs &1.1 \\ 
	& &  DV-Y &CG &abs &2.5 \\ 
	& &  DV-Y &SQR &rel &1.0 \\ 
	& &  DV-Y &CG &rel &2.5 \\ 
	& &  DV-Y &KNIFE &rel &1.1 \\ 
	& & DV-D &SQR &abs &1.0 \\ 
	& & DV-D &CG &abs &2.5 \\ 
	& & DV-D &KNIFE &abs &1.1 \\ 
	& & DV-D &KNIFE &rel &1.1 \\ 
	& & DV-D &CG &rel &2.5 \\ 
	& & DV-D &SQR &rel &1.0 \\ 
	\hline\multirow{10}{*}{boston} & \multirow{10}{*}{506} & DV-Y &KNIFE &abs &1.9 \\ 
	& &  DV-Y &CG &abs &4.1 \\ 
	& &  DV-Y &CG &rel &4.1 \\ 
	& &  DV-Y &KNIFE &rel &1.9 \\ 
	& & DV-D &SQR &abs &1.7 \\ 
	& & DV-D &CG &abs &4.1 \\ 
	& & DV-D &KNIFE &abs &1.8 \\ 
	& & DV-D &KNIFE &rel &1.8 \\ 
	& & DV-D &CG &rel &4.1 \\ 
	& & DV-D &SQR &rel &1.7 \\ 
	\hline\multirow{10}{*}{energy} & \multirow{10}{*}{768} & DV-Y &KNIFE &abs &2.7 \\ 
	& &  DV-Y &CG &abs &6.2 \\ 
	& &  DV-Y &CG &rel &6.3 \\ 
	& &  DV-Y &KNIFE &rel &2.7 \\ 
	& & DV-D &SQR &abs &2.6 \\ 
	& & DV-D &CG &abs &6.2 \\ 
	& & DV-D &KNIFE &abs &2.7 \\ 
	& & DV-D &KNIFE &rel &2.7 \\ 
	& & DV-D &CG &rel &6.2 \\ 
	& & DV-D &SQR &rel &2.6 \\ 
	\hline\multirow{10}{*}{concrete} & \multirow{10}{*}{1030} & DV-Y &KNIFE &abs &3.6 \\ 
	& &  DV-Y &CG &abs &8.3 \\ 
	& &  DV-Y &CG &rel &8.3 \\ 
	& &  DV-Y &KNIFE &rel &3.6 \\ 
	& & DV-D &SQR &abs &3.4 \\ 
	& & DV-D &CG &abs &8.3 \\ 
	& & DV-D &KNIFE &abs &3.6 \\ 
	& & DV-D &KNIFE &rel &3.6 \\ 
	& & DV-D &CG &rel &8.3 \\ 
	& & DV-D &SQR &rel &3.4 \\ 
	\hline\multirow{10}{*}{wine} & \multirow{10}{*}{4898} & DV-Y &KNIFE &abs &5.5 \\ 
	& &  DV-Y &SQR &abs &5.3 \\ 
	& &  DV-Y &CG &abs &13.0 \\ 
	& &  DV-Y &KNIFE &rel &5.5 \\ 
	& &  DV-Y &CG &rel &13.6 \\ 
	& &  DV-Y &SQR &rel &5.3 \\ 
	& & DV-D &SQR &abs &5.3 \\ 
	& & DV-D &CG &abs &12.8 \\ 
	& & DV-D &KNIFE &abs &5.5 \\ 
	& & DV-D &CG &rel &12.8 \\ 
	& & DV-D &SQR &rel &5.3 \\ 
	& & DV-D &KNIFE &rel &5.5 \\ 
	\hline\multirow{10}{*}{kin8nm} & \multirow{10}{*}{8192} & DV-Y &KNIFE &abs &28.9 \\ 
	& &  DV-Y &CG &abs &66.1 \\ 
	& &  DV-Y &CG &rel &66.1 \\ 
	& &  DV-Y &KNIFE &rel &28.7 \\ 
	& & DV-D &SQR &abs &27.2 \\ 
	& & DV-D &CG &abs &65.1 \\ 
	& & DV-D &KNIFE &abs &28.0 \\ 
	& & DV-D &KNIFE &rel &28.4 \\ 
	& & DV-D &CG &rel &65.4 \\ 
	& & DV-D &SQR &rel &27.1 \\ 
	\hline\multirow{10}{*}{power} & \multirow{10}{*}{9568} & DV-Y &KNIFE &abs &33.7 \\ 
	& &  DV-Y &CG &abs &76.9 \\ 
	& &  DV-Y &CG &rel &77.0 \\ 
	& &  DV-Y &KNIFE &rel &33.8 \\ 
	& & DV-D &SQR &abs &31.6 \\ 
	& & DV-D &CG &abs &76.5 \\ 
	& & DV-D &KNIFE &abs &33.0 \\ 
	& & DV-D &KNIFE &rel &33.1 \\ 
	& & DV-D &CG &rel &77.0 \\ 
	& & DV-D &SQR &rel &31.6 \\ 
	\hline\multirow{10}{*}{naval} & \multirow{10}{*}{11934} & DV-Y &KNIFE &abs &41.6 \\ 
	& &  DV-Y &CG &abs &96.5 \\ 
	& &  DV-Y &CG &rel &96.1 \\ 
	& &  DV-Y &KNIFE &rel &42.2 \\ 
	& & DV-D &SQR &abs &39.6 \\ 
	& & DV-D &CG &abs &95.3 \\ 
	& & DV-D &KNIFE &abs &41.7 \\ 
	& & DV-D &KNIFE &rel &41.7 \\ 
	& & DV-D &CG &rel &95.3 \\ 
	& & DV-D &SQR &rel &39.5 \\ 
	\hline\end{longtable}
}


%% file: tablesCF/small_rel_AUROC_base.txt
\begin{tabular}{|c|c|ccc|} \hline\multirow{14}{*}{\rotatebox[origin=c]{90}{\footnotesize \textbf{concrete}}}
& \cellcolor{tbl1} $\bm{\eps}$ \textbf{[\%]} & \cellcolor{tbl1} \textbf{10}\%  & \cellcolor{tbl1} \textbf{15}\%  & \cellcolor{tbl1} \textbf{20}\%  \\ \hhline{~----}
& \textbf{Avg. \% bad} \cellcolor{tbl1} & \cellcolor{tbl1}35.4 [6.2] & \cellcolor{tbl1}22.1 [5.1] & \cellcolor{tbl1}13.9 [3.4] \\ \hhline{~----}
& B1-SQR& 71.3 [5.3]& 74.8 [5.1]& 74.6 [6.3]\\ 
& B2-SQR& 66.5 [4.3]& 65.9 [3.8]& 62.9 [6.7]\\ 
& B1-KNIFE& 62.6 [4.2]& 63.2 [7.4]& 66.0 [6.1]\\ 
& B2-KNIFE& 61.4 [3.4]& 63.8 [5.1]& 64.9 [8.4]\\ 
& B1-CG& 62.9 [5.8]& 66.9 [8.0]& 70.6 [7.7]\\ 
& B2-CG& 67.2 [5.6]& 69.6 [8.2]& 71.3 [11.4]\\
\hline
\end{tabular}

%% file: tablesCF/small_rel_AUROC_CF.txt
\begin{tabular}{|c|c|ccc|} \hline\multirow{14}{*}{\rotatebox[origin=c]{90}{\footnotesize \textbf{concrete}}}
& \cellcolor{tbl1} $\bm{\eps}$ \textbf{[\%]} & \cellcolor{tbl1} \textbf{10}\%  & \cellcolor{tbl1} \textbf{15}\%  & \cellcolor{tbl1} \textbf{20}\%  \\ \hhline{~----}
& \textbf{Avg. \% bad} \cellcolor{tbl1} & \cellcolor{tbl1}35.4 [6.2] & \cellcolor{tbl1}22.1 [5.1] & \cellcolor{tbl1}13.9 [3.4] \\ \hhline{~----}
& B1-SQR-CF& 71.1 [5.5]& 74.8 [5.2]& 76.4 [5.0]\\
& B2-SQR-CF& 67.6 [4.4]& 69.1 [6.5]& 70.7 [8.1]\\
& B1-KNIFE-CF& 62.8 [3.5]& 64.6 [7.2]& 66.0 [5.7]\\
& B2-KNIFE-CF& 59.2 [4.2]& 62.7 [5.1]& 63.7 [8.5]\\
& B1-CG-CF& 62.7 [6.6]& 66.9 [7.9]& 70.3 [8.1]\\
& B2-CG-CF& 67.4 [6.1]& 69.5 [8.0]& 71.1 [10.7]\\
\hline
\end{tabular}

%% file: tablesCF/small_rel_FPR_base.txt
\begin{tabular}{|c|c|ccc|} \hline\multirow{13}{*}{\rotatebox[origin=c]{90}{\footnotesize \textbf{concrete}}}
& \cellcolor{tbl1} $\bm{\eps}$ \textbf{[\%]} & \cellcolor{tbl1} \textbf{10}\%  & \cellcolor{tbl1} \textbf{15}\%  & \cellcolor{tbl1} \textbf{20}\%  \\ \hhline{~----}
& B1-SQR& \textbf{0.63} [0.11]& 0.62 [0.21]& 0.76 [0.32]\\ 
& B2-SQR& 0.86 [0.15]& 0.95 [0.16]& 1.00 [0.00]\\ 
& B1-KNIFE& 0.76 [0.08]& 0.79 [0.11]& 0.73 [0.10]\\ 
& B2-KNIFE& 0.75 [0.04]& 0.73 [0.08]& 0.67 [0.12]\\ 
& B1-CG& 0.80 [0.04]& 0.69 [0.14]& 0.63 [0.19]\\ 
& B2-CG& 0.73 [0.09]& 0.64 [0.17]& 0.62 [0.23]\\ 
& DV-Y-SQR \cellcolor{tbl2}& \textbf{0.63} [0.11]\cellcolor{tbl2}& \textbf{0.56} [0.09]\cellcolor{tbl2}& \textbf{0.47} [0.14]\cellcolor{tbl2}\\
\hline
\end{tabular}

%% file: tablesCF/small_rel_FPR_CF.txt
\begin{tabular}{|c|c|ccc|} \hline\multirow{13}{*}{\rotatebox[origin=c]{90}{\footnotesize \textbf{concrete}}}
& \cellcolor{tbl1} $\bm{\eps}$ \textbf{[\%]} & \cellcolor{tbl1} \textbf{10}\%  & \cellcolor{tbl1} \textbf{15}\%  & \cellcolor{tbl1} \textbf{20}\%  \\ \hhline{~----}
& B1-SQR-CF& 0.66 [0.11]& 0.59 [0.13]& 0.50 [0.17]\\
& B2-SQR-CF& 0.74 [0.13]& 0.71 [0.11]& 0.66 [0.19]\\
& B1-KNIFE-CF& 0.76 [0.07]& 0.76 [0.09]& 0.70 [0.14]\\
& B2-KNIFE-CF& 0.79 [0.05]& 0.76 [0.07]& 0.71 [0.10]\\
& B1-CG-CF& 0.80 [0.06]& 0.70 [0.15]& 0.64 [0.21]\\
& B2-CG-CF& 0.74 [0.12]& 0.67 [0.16]& 0.65 [0.21]\\
& & & & \\
\hline
\end{tabular}

%% file: tablesCF/small_abs_AUROC_base.txt
\begin{tabular}{|c|c|ccc|} \hline\multirow{14}{*}{\rotatebox[origin=c]{90}{\footnotesize \textbf{concrete}}}
& \cellcolor{tbl1} $\bm{\eps}$ & \cellcolor{tbl1} \textbf{0.2}  & \cellcolor{tbl1} \textbf{0.5}  & \cellcolor{tbl1} \textbf{0.6}  \\ \hhline{~----}
& \textbf{Avg. \% bad} \cellcolor{tbl1} & \cellcolor{tbl1}36.7 [5.6] & \cellcolor{tbl1}9.6 [2.8] & \cellcolor{tbl1}6.4 [2.0] \\ \hhline{~----}
& B1-SQR& 73.1 [3.8]& 81.1 [8.3]& 76.3 [14.2]\\ 
& B2-SQR& 69.8 [7.3]& 59.2 [9.0]& 57.3 [9.3]\\ 
& B1-KNIFE& 63.8 [7.4]& 68.7 [5.9]& 71.9 [9.6]\\ 
& B2-KNIFE& 67.7 [6.9]& 76.9 [8.9]& 80.9 [12.2]\\ 
& B1-CG& 66.6 [4.7]& 77.6 [5.6]& 79.8 [7.2]\\ 
& B2-CG& 69.8 [7.1]& 78.7 [9.6]& 81.5 [9.1]\\
\hline
\end{tabular}

%% file: tablesCF/small_abs_AUROC_CF.txt
\begin{tabular}{|c|c|ccc|} \hline\multirow{14}{*}{\rotatebox[origin=c]{90}{\footnotesize \textbf{concrete}}}
& \cellcolor{tbl1} $\bm{\eps}$ & \cellcolor{tbl1} \textbf{0.2}  & \cellcolor{tbl1} \textbf{0.5}  & \cellcolor{tbl1} \textbf{0.6}  \\ \hhline{~----}
& \textbf{Avg. \% bad} \cellcolor{tbl1} & \cellcolor{tbl1}36.7 [5.6] & \cellcolor{tbl1}9.6 [2.8] & \cellcolor{tbl1}6.4 [2.0] \\ \hhline{~----}
& B1-SQR-CF& 73.3 [3.3]& 81.4 [10.3]& 80.0 [13.1]\\
& B2-SQR-CF& 70.1 [6.2]& 76.1 [9.4]& 72.4 [10.6]\\
& B1-KNIFE-CF& 64.7 [6.9]& 69.9 [5.8]& 72.8 [8.8]\\
& B2-KNIFE-CF& 67.8 [7.4]& 76.6 [9.3]& 77.3 [14.5]\\
& B1-CG-CF& 66.9 [4.8]& 77.8 [5.5]& 79.3 [7.7]\\
& B2-CG-CF& 70.0 [6.9]& 74.5 [13.7]& 74.2 [16.0]\\
\hline\end{tabular}

%% file: tablesCF/small_abs_FPR_base.txt
\begin{tabular}{|c|c|ccc|} \hline\multirow{13}{*}{\rotatebox[origin=c]{90}{\footnotesize \textbf{concrete}}}
& \cellcolor{tbl1} $\bm{\eps}$ & \cellcolor{tbl1} \textbf{0.2}  & \cellcolor{tbl1} \textbf{0.5}  & \cellcolor{tbl1} \textbf{0.6}  \\ \hhline{~----}
& B1-SQR& \textbf{0.67} [0.10]& 0.77 [0.39]& 0.90 [0.30]\\ 
& B2-SQR& 0.74 [0.13]& 1.00 [0.00]& 1.00 [0.00]\\ 
& B1-KNIFE& 0.76 [0.08]& 0.68 [0.16]& 0.61 [0.21]\\ 
& B2-KNIFE& 0.73 [0.13]& 0.54 [0.18]& \textbf{0.45} [0.27]\\ 
& B1-CG& 0.69 [0.10]& \textbf{0.44} [0.13]& 0.49 [0.18]\\ 
& B2-CG& \textbf{0.68} [0.12]& 0.50 [0.22]& 0.49 [0.23]\\ 
& DV-Y-SQR \cellcolor{tbl2}& \textbf{0.67} [0.10]\cellcolor{tbl2}& \textbf{0.43} [0.21]\cellcolor{tbl2}& 0.51 [0.22]\cellcolor{tbl2}\\
\hline
\end{tabular}

%% file: tablesCF/small_abs_FPR_CF.txt
\begin{tabular}{|c|c|ccc|} \hline\multirow{13}{*}{\rotatebox[origin=c]{90}{\footnotesize \textbf{concrete}}}
& \cellcolor{tbl1} $\bm{\eps}$ & \cellcolor{tbl1} \textbf{0.2}  & \cellcolor{tbl1} \textbf{0.5}  & \cellcolor{tbl1} \textbf{0.6}  \\ \hhline{~----}
& B1-SQR-CF& 0.69 [0.13]& 0.44 [0.25]& 0.61 [0.30]\\
& B2-SQR-CF& 0.67 [0.12]& 0.79 [0.26]& 0.87 [0.27]\\
& B1-KNIFE-CF& 0.74 [0.05]& 0.68 [0.15]& 0.64 [0.21]\\
& B2-KNIFE-CF& 0.74 [0.12]& 0.66 [0.28]& 0.64 [0.40]\\
& B1-CG-CF& 0.66 [0.08]& 0.47 [0.14]& 0.51 [0.19]\\
& B2-CG-CF& 0.69 [0.11]& 0.70 [0.33]& 0.79 [0.32]\\
& & & & \\
\hline\end{tabular}

%% file: tablesCF/large_rel_AUROC_base.txt
\begin{tabular}{|c|c|ccc|} \hline\multirow{14}{*}{\rotatebox[origin=c]{90}{\footnotesize \textbf{naval}}}
& \cellcolor{tbl1} $\bm{\eps}$ \textbf{[\%]} & \cellcolor{tbl1} \textbf{0.15}\%  & \cellcolor{tbl1} \textbf{0.175}\%  & \cellcolor{tbl1} \textbf{0.2}\%  \\ \hhline{~----}
& \textbf{Avg. \% bad} \cellcolor{tbl1} & \cellcolor{tbl1}1.5 [0.6] & \cellcolor{tbl1}0.8 [0.4] & \cellcolor{tbl1}0.5 [0.3] \\ \hhline{~----}
& B1-SQR& 83.7 [5.0]& 90.4 [6.1]& 89.7 [11.2]\\ 
& B2-SQR& 94.8 [4.1]& 87.6 [7.2]& 90.2 [9.7]\\ 
& B1-KNIFE& 59.0 [10.0]& 68.7 [13.5]& 62.1 [19.8]\\ 
& B2-KNIFE& 92.4 [3.8]& 95.1 [3.2]& 96.7 [2.7]\\ 
& B1-CG& 89.3 [3.1]& 91.2 [2.3]& 89.7 [3.2]\\ 
& B2-CG& 93.0 [1.6]& 94.1 [4.1]& 95.2 [4.6]\\
\hline
\end{tabular}

%% file: tablesCF/large_rel_AUROC_CF.txt
\begin{tabular}{|c|c|ccc|} \hline\multirow{14}{*}{\rotatebox[origin=c]{90}{\footnotesize \textbf{naval}}}
& \cellcolor{tbl1} $\bm{\eps}$ \textbf{[\%]} & \cellcolor{tbl1} \textbf{0.15}\%  & \cellcolor{tbl1} \textbf{0.175}\%  & \cellcolor{tbl1} \textbf{0.2}\%  \\ \hhline{~----}
& \textbf{Avg. \% bad} \cellcolor{tbl1} & \cellcolor{tbl1}1.5 [0.6] & \cellcolor{tbl1}0.8 [0.4] & \cellcolor{tbl1}0.5 [0.3] \\ \hhline{~----}
& B1-SQR-CF& 83.3 [6.5]& 90.1 [6.8]& 90.7 [8.7]\\
& B2-SQR-CF& 97.2 [1.6]& 94.8 [5.9]& 93.9 [7.1]\\
& B1-KNIFE-CF& 67.8 [9.2]& 72.8 [13.7]& 70.1 [15.1]\\
& B2-KNIFE-CF& 91.0 [5.2]& 89.4 [11.7]& 89.4 [13.7]\\
& B1-CG-CF& 88.6 [4.4]& 90.3 [2.6]& 88.6 [2.8]\\
& B2-CG-CF& 92.4 [1.9]& 92.0 [8.2]& 91.4 [10.5]\\
\hline
\hline\end{tabular}

%% file: tablesCF/large_rel_FPR_base.txt
\begin{tabular}{|c|c|ccc|} \hline\multirow{13}{*}{\rotatebox[origin=c]{90}{\footnotesize \textbf{naval}}}
& \cellcolor{tbl1} $\bm{\eps}$ \textbf{[\%]} & \cellcolor{tbl1} \textbf{0.15}\%  & \cellcolor{tbl1} \textbf{0.175}\%  & \cellcolor{tbl1} \textbf{0.2}\%  \\ \hhline{~----}
& B1-SQR& 0.43 [0.13]& 0.26 [0.17]& 0.38 [0.44]\\ 
& B2-SQR& 0.33 [0.46]& 0.80 [0.41]& 0.60 [0.51]\\ 
& B1-KNIFE& 0.88 [0.11]& 0.63 [0.28]& 0.70 [0.26]\\ 
& B2-KNIFE& 0.22 [0.13]& 0.12 [0.07]& \textbf{0.09} [0.08]\\ 
& B1-CG& 0.23 [0.09]& 0.14 [0.04]& 0.15 [0.05]\\ 
& B2-CG& 0.16 [0.03]& 0.14 [0.07]& 0.11 [0.09]\\ 
& DV-Y-SQR \cellcolor{tbl2}& 0.45 [0.24]\cellcolor{tbl2}& 0.29 [0.25]\cellcolor{tbl2}& 0.20 [0.20]\cellcolor{tbl2}\\
\hline
\end{tabular}

%% file: tablesCF/large_rel_FPR_CF.txt
\begin{tabular}{|c|c|ccc|} \hline\multirow{13}{*}{\rotatebox[origin=c]{90}{\footnotesize \textbf{naval}}}
& \cellcolor{tbl1} $\bm{\eps}$ \textbf{[\%]} & \cellcolor{tbl1} \textbf{0.15}\%  & \cellcolor{tbl1} \textbf{0.175}\%  & \cellcolor{tbl1} \textbf{0.2}\%  \\ \hhline{~----}
& B1-SQR-CF& 0.49 [0.21]& 0.30 [0.23]& 0.35 [0.39]\\
& B2-SQR-CF& \textbf{0.10} [0.08]& 0.24 [0.40]& 0.41 [0.51]\\
& B1-KNIFE-CF& 0.73 [0.19]& 0.54 [0.33]& 0.52 [0.29]\\
& B2-KNIFE-CF& 0.31 [0.29]& 0.33 [0.37]& 0.37 [0.45]\\
& B1-CG-CF& 0.23 [0.11]& 0.15 [0.04]& 0.16 [0.05]\\
& B2-CG-CF& 0.16 [0.04]& 0.29 [0.38]& 0.35 [0.45]\\
& & & & \\
\hline\end{tabular}

%% file: tablesCF/large_abs_AUROC_base.txt
\begin{tabular}{|c|c|ccc|} \hline\multirow{14}{*}{\rotatebox[origin=c]{90}{\footnotesize \textbf{naval}}}
& \cellcolor{tbl1} $\bm{\eps}$ & \cellcolor{tbl1} \textbf{0.175}  & \cellcolor{tbl1} \textbf{0.2}  & \cellcolor{tbl1} \textbf{0.225}  \\ \hhline{~----}
& \textbf{Avg. \% bad} \cellcolor{tbl1} & \cellcolor{tbl1}2.6 [0.9] & \cellcolor{tbl1}1.4 [0.7] & \cellcolor{tbl1}0.9 [0.4] \\ \hhline{~----}
& B1-SQR& 78.2 [8.5]& 84.7 [5.4]& 89.0 [5.7]\\ 
& B2-SQR& 93.9 [3.8]& 93.9 [4.2]& 88.8 [7.2]\\ 
& B1-KNIFE& 56.9 [8.4]& 59.1 [10.7]& 66.4 [13.4]\\ 
& B2-KNIFE& 91.6 [3.5]& 93.4 [3.6]& 95.2 [2.7]\\ 
& B1-CG& 88.3 [3.8]& 89.2 [3.0]& 90.5 [2.0]\\ 
& B2-CG& 91.7 [1.1]& 93.0 [2.0]& 94.4 [2.6]\\
\hline
\end{tabular}

%% file: tablesCF/large_abs_AUROC_CF.txt
\begin{tabular}{|c|c|ccc|} \hline\multirow{14}{*}{\rotatebox[origin=c]{90}{\footnotesize \textbf{naval}}}
& \cellcolor{tbl1} $\bm{\eps}$ & \cellcolor{tbl1} \textbf{0.175}  & \cellcolor{tbl1} \textbf{0.2}  & \cellcolor{tbl1} \textbf{0.225}  \\ \hhline{~----}
& \textbf{Avg. \% bad} \cellcolor{tbl1} & \cellcolor{tbl1}2.6 [0.9] & \cellcolor{tbl1}1.4 [0.7] & \cellcolor{tbl1}0.9 [0.4] \\ \hhline{~----}
& B1-SQR-CF& 77.9 [7.9]& 84.3 [7.2]& 88.3 [5.9]\\
& B2-SQR-CF& 96.5 [1.5]& \textbf{97.3} [1.8]& 96.1 [4.7]\\
& B1-KNIFE-CF& 67.3 [8.6]& 67.4 [8.7]& 70.0 [11.1]\\
& B2-KNIFE-CF& 86.5 [7.5]& 87.9 [6.4]& 85.3 [10.8]\\
& B1-CG-CF& 87.9 [5.1]& 88.6 [4.4]& 89.5 [2.0]\\
& B2-CG-CF& 91.1 [1.6]& 92.0 [2.5]& 93.3 [3.9]\\
\hline
\end{tabular}

%% file: tablesCF/large_abs_FPR_base.txt
\begin{tabular}{|c|c|ccc|} \hline\multirow{13}{*}{\rotatebox[origin=c]{90}{\footnotesize \textbf{naval}}}
& \cellcolor{tbl1} $\bm{\eps}$ & \cellcolor{tbl1} \textbf{0.175}  & \cellcolor{tbl1} \textbf{0.2}  & \cellcolor{tbl1} \textbf{0.225}  \\ \hhline{~----}
& B1-SQR& 0.58 [0.18]& 0.45 [0.20]& 0.31 [0.12]\\ 
& B2-SQR& 0.45 [0.48]& 0.53 [0.50]& 0.80 [0.41]\\ 
& B1-KNIFE& 0.88 [0.06]& 0.87 [0.13]& 0.70 [0.25]\\ 
& B2-KNIFE& 0.22 [0.11]& 0.19 [0.13]& \textbf{0.12} [0.06]\\ 
& B1-CG& 0.25 [0.07]& 0.23 [0.09]& 0.18 [0.07]\\ 
& B2-CG& 0.20 [0.04]& 0.19 [0.05]& 0.15 [0.07]\\ 
& DV-Y-SQR \cellcolor{tbl2}& 0.53 [0.19]\cellcolor{tbl2}& 0.45 [0.26]\cellcolor{tbl2}& 0.30 [0.13]\cellcolor{tbl2}\\
\hline
\end{tabular}

%% file: tablesCF/large_abs_FPR_CF.txt
\begin{tabular}{|c|c|ccc|} \hline\multirow{13}{*}{\rotatebox[origin=c]{90}{\footnotesize \textbf{naval}}}
& \cellcolor{tbl1} $\bm{\eps}$ & \cellcolor{tbl1} \textbf{0.175}  & \cellcolor{tbl1} \textbf{0.2}  & \cellcolor{tbl1} \textbf{0.225}  \\ \hhline{~----}
& B1-SQR-CF& 0.63 [0.15]& 0.49 [0.26]& 0.38 [0.22]\\
& B2-SQR-CF& 0.12 [0.05]& \textbf{0.09} [0.04]& 0.25 [0.40]\\
& B1-KNIFE-CF& 0.72 [0.20]& 0.79 [0.16]& 0.66 [0.27]\\
& B2-KNIFE-CF& 0.29 [0.12]& 0.34 [0.27]& 0.44 [0.41]\\
& B1-CG-CF& 0.26 [0.09]& 0.24 [0.11]& 0.18 [0.06]\\
& B2-CG-CF& 0.21 [0.04]& 0.34 [0.35]& 0.23 [0.28]\\
& & & & \\
\hline\end{tabular}

%% file: tables2/small_abs.txt
\begin{table}[ht!] \small \centering  \caption{AUROC for the  absolute error discrepancy metric   for  the small  datasets. B1, B2 correspond to baseline Algorithms \ref{alg:baseline1} and \ref{alg:baseline2} resp. DV-Y and DV-D  correspond to the proposed Algorithms \ref{alg:div_discY} and \ref{alg:div_discD} resp.}
\vskip 0.15in
\begin{tabular}{|c|c|ccc|} \hline\multirow{14}{*}{\rotatebox[origin=c]{90}{\footnotesize \textbf{concrete}}}
& \cellcolor{tbl1} $\bm{\eps}$ & \cellcolor{tbl1} \textbf{0.2}  & \cellcolor{tbl1} \textbf{0.5}  & \cellcolor{tbl1} \textbf{0.6}  \\ \hhline{~----}
& \textbf{Avg. \% bad} \cellcolor{tbl1} & \cellcolor{tbl1}36.7 [5.6] & \cellcolor{tbl1}9.6 [2.8] & \cellcolor{tbl1}6.4 [2.0] \\ \hhline{~----}
& B1-SQR& 73.1 [3.8]& 81.1 [8.3]& 76.3 [14.2]\\ 
& B2-SQR& 69.8 [7.3]& 59.2 [9.0]& 57.3 [9.3]\\ 
& B1-KNIFE& 63.8 [7.4]& 68.7 [5.9]& 71.9 [9.6]\\ 
& B2-KNIFE& 67.7 [6.9]& 76.9 [8.9]& 80.9 [12.2]\\ 
& B1-CG& 66.6 [4.7]& 77.6 [5.6]& 79.8 [7.2]\\ 
& B2-CG& 69.8 [7.1]& 78.7 [9.6]& 81.5 [9.1]\\
\hline
\end{tabular}\end{table}

\begin{table}[ht!] \centering \small \caption{FPR  at TPR 90\% for the  absolute error discrepancy metric   for  the small  datasets. B1, B2 correspond to baseline Algorithms \ref{alg:baseline1} and \ref{alg:baseline2} resp. DV-Y and DV-D  correspond to the proposed Algorithms \ref{alg:div_discY} and \ref{alg:div_discD} resp.}
\vskip 0.15in
\begin{tabular}{|c|c|ccc|} \hline\multirow{13}{*}{\rotatebox[origin=c]{90}{\footnotesize \textbf{concrete}}}
& \cellcolor{tbl1} $\bm{\eps}$ & \cellcolor{tbl1} \textbf{0.2}  & \cellcolor{tbl1} \textbf{0.5}  & \cellcolor{tbl1} \textbf{0.6}  \\ \hhline{~----}
& B1-SQR& \textbf{0.67} [0.10]& 0.77 [0.39]& 0.90 [0.30]\\ 
& B2-SQR& 0.74 [0.13]& 1.00 [0.00]& 1.00 [0.00]\\ 
& B1-KNIFE& 0.76 [0.08]& 0.68 [0.16]& 0.61 [0.21]\\ 
& B2-KNIFE& 0.73 [0.13]& 0.54 [0.18]& \textbf{0.45} [0.27]\\ 
& B1-CG& 0.69 [0.10]& \textbf{0.44} [0.13]& 0.49 [0.18]\\ 
& B2-CG& \textbf{0.68} [0.12]& 0.50 [0.22]& 0.49 [0.23]\\ 
& B2-RIO& 0.73 [0.15]& 0.62 [0.22]& 0.55 [0.25]\\
\hline
\end{tabular}\end{table}

%% file: tables2/large_abs.txt
\begin{table}[ht!] \small \centering  \caption{AUROC for the  absolute error discrepancy metric   for  the large  datasets. B1, B2 correspond to baseline Algorithms \ref{alg:baseline1} and \ref{alg:baseline2} resp. DV-Y and DV-D  correspond to the proposed Algorithms \ref{alg:div_discY} and \ref{alg:div_discD} resp.}
\vskip 0.15in
\begin{tabular}{|c|c|ccc|} \hline\multirow{14}{*}{\rotatebox[origin=c]{90}{\footnotesize \textbf{naval}}}
& \cellcolor{tbl1} $\bm{\eps}$ & \cellcolor{tbl1} \textbf{0.175}  & \cellcolor{tbl1} \textbf{0.2}  & \cellcolor{tbl1} \textbf{0.225}  \\ \hhline{~----}
& \textbf{Avg. \% bad} \cellcolor{tbl1} & \cellcolor{tbl1}2.6 [0.9] & \cellcolor{tbl1}1.4 [0.7] & \cellcolor{tbl1}0.9 [0.4] \\ \hhline{~----}
& B1-SQR& 78.2 [8.5]& 84.7 [5.4]& 89.0 [5.7]\\ 
& B2-SQR& 93.9 [3.8]& 93.9 [4.2]& 88.8 [7.2]\\ 
& B1-KNIFE& 56.9 [8.4]& 59.1 [10.7]& 66.4 [13.4]\\ 
& B2-KNIFE& 91.6 [3.5]& 93.4 [3.6]& 95.2 [2.7]\\ 
& B1-CG& 88.3 [3.8]& 89.2 [3.0]& 90.5 [2.0]\\ 
& B2-CG& 91.7 [1.1]& 93.0 [2.0]& 94.4 [2.6]\\
\hline
\end{tabular}\end{table}

\begin{table}[ht!] \centering \small \caption{FPR  at TPR 90\% for the  absolute error discrepancy metric   for  the large  datasets. B1, B2 correspond to baseline Algorithms \ref{alg:baseline1} and \ref{alg:baseline2} resp. DV-Y and DV-D  correspond to the proposed Algorithms \ref{alg:div_discY} and \ref{alg:div_discD} resp.}
\vskip 0.15in
\begin{tabular}{|c|c|ccc|} \hline\multirow{13}{*}{\rotatebox[origin=c]{90}{\footnotesize \textbf{naval}}}
& \cellcolor{tbl1} $\bm{\eps}$ & \cellcolor{tbl1} \textbf{0.175}  & \cellcolor{tbl1} \textbf{0.2}  & \cellcolor{tbl1} \textbf{0.225}  \\ \hhline{~----}
& B1-SQR& 0.58 [0.18]& 0.45 [0.20]& 0.31 [0.12]\\ 
& B2-SQR& 0.45 [0.48]& 0.53 [0.50]& 0.80 [0.41]\\ 
& B1-KNIFE& 0.88 [0.06]& 0.87 [0.13]& 0.70 [0.25]\\ 
& B2-KNIFE& 0.22 [0.11]& 0.19 [0.13]& \textbf{0.12} [0.06]\\ 
& B1-CG& 0.25 [0.07]& 0.23 [0.09]& 0.18 [0.07]\\ 
& B2-CG& 0.20 [0.04]& 0.19 [0.05]& 0.15 [0.07]\\ 
& B2-RIO& 0.25 [0.07]& 0.27 [0.10]& 0.24 [0.12]\\
\hline
\end{tabular}\end{table}

%% file: tables2/small_rel.txt
\begin{table}[ht!] \small \centering  \caption{AUROC for the  relative error discrepancy metric   for  the small  datasets. B1, B2 correspond to baseline Algorithms \ref{alg:baseline1} and \ref{alg:baseline2} resp. DV-Y and DV-D  correspond to the proposed Algorithms \ref{alg:div_discY} and \ref{alg:div_discD} resp.}\vskip 0.15in
\begin{tabular}{|c|c|ccc|} \hline\multirow{14}{*}{\rotatebox[origin=c]{90}{\footnotesize \textbf{concrete}}}
& \cellcolor{tbl1} $\bm{\eps}$ \textbf{[\%]} & \cellcolor{tbl1} \textbf{10}\%  & \cellcolor{tbl1} \textbf{15}\%  & \cellcolor{tbl1} \textbf{20}\%  \\ \hhline{~----}
& \textbf{Avg. \% bad} \cellcolor{tbl1} & \cellcolor{tbl1}35.4 [6.2] & \cellcolor{tbl1}22.1 [5.1] & \cellcolor{tbl1}13.9 [3.4] \\ \hhline{~----}
& B1-SQR& 71.3 [5.3]& 74.8 [5.1]& 74.6 [6.3]\\ 
& B2-SQR& 66.5 [4.3]& 65.9 [3.8]& 62.9 [6.7]\\ 
& B1-KNIFE& 62.6 [4.2]& 63.2 [7.4]& 66.0 [6.1]\\ 
& B2-KNIFE& 61.4 [3.4]& 63.8 [5.1]& 64.9 [8.4]\\ 
& B1-CG& 62.9 [5.8]& 66.9 [8.0]& 70.6 [7.7]\\ 
& B2-CG& 67.2 [5.6]& 69.6 [8.2]& 71.3 [11.4]\\
\hline
\end{tabular}\end{table}

\begin{table}[ht!] \centering \small \caption{FPR  at TPR 90\% for the  relative error discrepancy metric   for  the small  datasets. B1, B2 correspond to baseline Algorithms \ref{alg:baseline1} and \ref{alg:baseline2} resp. DV-Y and DV-D  correspond to the proposed Algorithms \ref{alg:div_discY} and \ref{alg:div_discD} resp.}\vskip 0.15in
\begin{tabular}{|c|c|ccc|} \hline\multirow{13}{*}{\rotatebox[origin=c]{90}{\footnotesize \textbf{concrete}}}
& \cellcolor{tbl1} $\bm{\eps}$ \textbf{[\%]} & \cellcolor{tbl1} \textbf{10}\%  & \cellcolor{tbl1} \textbf{15}\%  & \cellcolor{tbl1} \textbf{20}\%  \\ \hhline{~----}
& B1-SQR& \textbf{0.63} [0.11]& 0.62 [0.21]& 0.76 [0.32]\\ 
& B2-SQR& 0.86 [0.15]& 0.95 [0.16]& 1.00 [0.00]\\ 
& B1-KNIFE& 0.76 [0.08]& 0.79 [0.11]& 0.73 [0.10]\\ 
& B2-KNIFE& 0.75 [0.04]& 0.73 [0.08]& 0.67 [0.12]\\ 
& B1-CG& 0.80 [0.04]& 0.69 [0.14]& 0.63 [0.19]\\ 
& B2-CG& 0.73 [0.09]& 0.64 [0.17]& 0.62 [0.23]\\ 
& B2-RIO& 0.71 [0.13]& 0.72 [0.14]& 0.70 [0.24]\\
\hline
\end{tabular}\end{table}

%% file: tables2/large_rel.txt
\begin{table}[ht!] \small \centering  \caption{AUROC for the  relative error discrepancy metric   for  the large  datasets. B1, B2 correspond to baseline Algorithms \ref{alg:baseline1} and \ref{alg:baseline2} resp. DV-Y and DV-D  correspond to the proposed Algorithms \ref{alg:div_discY} and \ref{alg:div_discD} resp.}\vskip 0.15in
\begin{tabular}{|c|c|ccc|} \hline\multirow{14}{*}{\rotatebox[origin=c]{90}{\footnotesize \textbf{naval}}}
& \cellcolor{tbl1} $\bm{\eps}$ \textbf{[\%]} & \cellcolor{tbl1} \textbf{0.15}\%  & \cellcolor{tbl1} \textbf{0.175}\%  & \cellcolor{tbl1} \textbf{0.2}\%  \\ \hhline{~----}
& \textbf{Avg. \% bad} \cellcolor{tbl1} & \cellcolor{tbl1}1.5 [0.6] & \cellcolor{tbl1}0.8 [0.4] & \cellcolor{tbl1}0.5 [0.3] \\ \hhline{~----}
& B1-SQR& 83.7 [5.0]& 90.4 [6.1]& 89.7 [11.2]\\ 
& B2-SQR& 94.8 [4.1]& 87.6 [7.2]& 90.2 [9.7]\\ 
& B1-KNIFE& 59.0 [10.0]& 68.7 [13.5]& 62.1 [19.8]\\ 
& B2-KNIFE& 92.4 [3.8]& 95.1 [3.2]& 96.7 [2.7]\\ 
& B1-CG& 89.3 [3.1]& 91.2 [2.3]& 89.7 [3.2]\\ 
& B2-CG& 93.0 [1.6]& 94.1 [4.1]& 95.2 [4.6]\\
\hline
\end{tabular}\end{table}

\begin{table}[ht!] \centering \small \caption{FPR  at TPR 90\% for the  relative error discrepancy metric   for  the large  datasets. B1, B2 correspond to baseline Algorithms \ref{alg:baseline1} and \ref{alg:baseline2} resp. DV-Y and DV-D  correspond to the proposed Algorithms \ref{alg:div_discY} and \ref{alg:div_discD} resp.}\vskip 0.15in
\begin{tabular}{|c|c|ccc|} \hline\multirow{13}{*}{\rotatebox[origin=c]{90}{\footnotesize \textbf{naval}}}
& \cellcolor{tbl1} $\bm{\eps}$ \textbf{[\%]} & \cellcolor{tbl1} \textbf{0.15}\%  & \cellcolor{tbl1} \textbf{0.175}\%  & \cellcolor{tbl1} \textbf{0.2}\%  \\ \hhline{~----}
& B1-SQR& 0.43 [0.13]& 0.26 [0.17]& 0.38 [0.44]\\ 
& B2-SQR& 0.33 [0.46]& 0.80 [0.41]& 0.60 [0.51]\\ 
& B1-KNIFE& 0.88 [0.11]& 0.63 [0.28]& 0.70 [0.26]\\ 
& B2-KNIFE& 0.22 [0.13]& 0.12 [0.07]& \textbf{0.09} [0.08]\\ 
& B1-CG& 0.23 [0.09]& 0.14 [0.04]& 0.15 [0.05]\\ 
& B2-CG& 0.16 [0.03]& 0.14 [0.07]& 0.11 [0.09]\\ 
& B2-RIO& 0.25 [0.08]& 0.24 [0.16]& 0.22 [0.14]\\
\hline
\end{tabular}\end{table}